\documentclass[10pt,twocolumn,letterpaper]{article}

\usepackage{cvpr}              

\usepackage[accsupp]{axessibility}

\definecolor{cvprblue}{rgb}{0.21,0.49,0.74}
\usepackage[pagebackref,breaklinks,colorlinks,allcolors=cvprblue]{hyperref}

\usepackage[table]{xcolor}
\usepackage{multirow}
\usepackage{makecell}
\usepackage{booktabs}
\usepackage{arydshln}
\usepackage{bm}

\def\paperID{12586} 
\def\confName{CVPR}
\def\confYear{2026}

\title{Enhancing Mixture-of-Experts Specialization via Cluster-Aware Upcycling}

\author{
Sanghyeok Chu\footnotemark[1]~~$^{1,2}$ \qquad 
Pyunghwan Ahn\footnotemark[2]~~$^{2}$ \qquad 
Gwangmo Song$^{2}$ \qquad 
SeungHwan Kim$^{2}$ \qquad \\
Honglak Lee$^{2,3}$ \qquad
Bohyung Han\footnotemark[2]~~$^{1,4}$  \\
$^1$ECE \& $^4$IPAI, Seoul National University \qquad
$^2$LG AI Research \qquad
$^3$University of Michigan \\
{\tt\small
\{sanghyeok.chu, bhhan\}@snu.ac.kr \quad
\{p.ahn, gwangmo.song, sh.kim, honglak\}@lgresearch.ai
}
}

\begin{document}
\maketitle
\renewcommand{\thefootnote}{\fnsymbol{footnote}}
\footnotetext[1]{Work done during an internship at LG AI Research.}
\footnotetext[2]{Corresponding authors.}
\renewcommand{\thefootnote}{\arabic{footnote}}
\begin{abstract}
Sparse Upcycling provides an efficient way to initialize a Mixture-of-Experts (MoE) model from pretrained dense weights instead of training from scratch.
However, since all experts start from identical weights and the router is randomly initialized, the model suffers from expert symmetry and limited early specialization.
We propose Cluster-aware Upcycling, a strategy that incorporates semantic structure into MoE initialization.
Our method first partitions the dense model’s input activations into semantic clusters. 
Each expert is then initialized using the subspace representations of its corresponding cluster via truncated SVD, while setting the router's initial weights to the cluster centroids.
This cluster-aware initialization breaks expert symmetry and encourages early specialization aligned with the {data distribution.}
Furthermore, we introduce an expert-ensemble self-distillation loss that stabilizes training by providing reliable routing guidance using an ensemble teacher.
When evaluated on CLIP ViT-B/32 and ViT-B/16, Cluster-aware Upcycling consistently outperforms existing methods across both zero-shot and few-shot benchmarks.
The proposed method also produces more diverse and disentangled expert representations, reduces inter-expert similarity, and leads to more confident routing behavior.
\vspace{-1mm}
\end{abstract}

\section{Introduction}
Scaling model size has been a reliable strategy for improving performance~\cite{kaplan2020scaling, li2023scaling, dubey2024llama, yang2025qwen3, comanici2025gemini}.
However, this scaling incurs a significant computational cost.
In standard dense architectures, where all parameters are activated for every input, training and inference costs grow linearly with model size.
Mixture-of-Experts (MoE) architectures offer a sparse alternative by activating only a subset of parameters for each input token~\cite{lepikhin2020gshard, fedus2022switch}.  
This conditional computation enables models to scale efficiently without a proportional increase in computational cost, particularly during inference.

However, training MoE models from scratch remains prohibitively expensive. 
Therefore, Sparse Upcycling~\cite{komatsuzaki2023sparse} has emerged as an effective warm-start strategy that initializes an MoE model by reusing pretrained dense weights.
This approach preserves the dense model’s functionality at initialization, leading to faster convergence.
However, initializing all experts with identical weights alongside a randomly initialized router inherently introduces expert symmetry; consequently, the model lacks a meaningful basis for early specialization.
Prior work has attempted to break this symmetry through noise injection~\cite{komatsuzaki2023sparse, chi2022representation, team2024qwen2, muennighoff2025olmoe}, but these approaches often yield marginal performance gains. 
Drop-Upcycling~\cite{nakamura2025dropupcycling} partially reinitializes expert parameters to encourage diversity, but this inevitably disrupts the pretrained representation space and undermines the core benefits of upcycling.
Other methods fine-tune experts on manually defined domains to induce specialization~\cite{sukhbaatarbranch, gritsch2024nexus}, but this requires explicit domain partitioning and additional fine-tuning stages, which fail to scale effectively to models with a large number of experts.
\begin{figure}[t]
    \centering    \includegraphics[width=0.975\columnwidth]{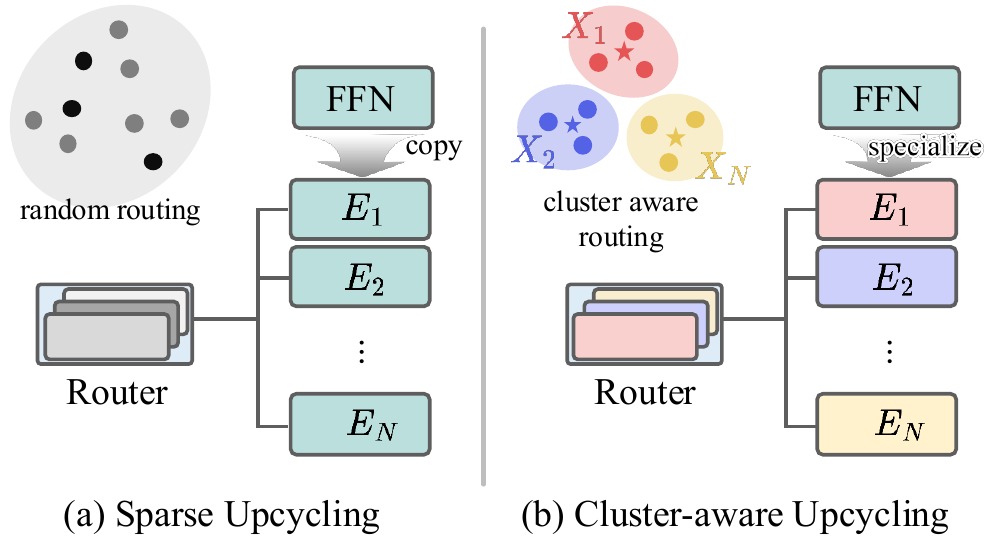}
    \caption{Comparison of Sparse Upcycling and Cluster-aware Upcycling.
    Unlike Sparse Upcycling, which inherently introduces expert symmetry, our method leverages semantic structure to initialize both experts and the router, promoting expert specialization.
    }
    \vspace{-2mm}
    \label{fig:overview}
\end{figure}
%

The key insight of our work is that the representations of a pretrained dense model already contain semantic information that can effectively guide the initialization of both experts and router parameters.
Rather than treating the dense model as a monolithic entity to be replicated, we leverage its activation manifold to initialize experts into distinct subspaces while faithfully preserving pretrained knowledge.

Building on this insight, we propose Cluster-aware Upcycling, a strategy that incorporates semantic structure into MoE initialization.
We utilize spherical $k$-means clustering to partition the activation space based on cosine similarity, which directly aligns with the routing mechanism.
Each expert is initialized to capture the subspace of its corresponding cluster via truncated SVD.
The router is initialized using the cluster centroids, ensuring that early routing decisions align with the underlying semantic structure of the data.
This cluster-aware initialization breaks expert symmetry and provides the router with a meaningful prior for expert specialization from the onset of training.

In addition to the initialization strategy, we propose the expert-ensemble self-distillation (EESD) loss to further enhance training.
Tokens with near-uniform routing probabilities often lack clear expert assignments, which can hinder the development of expert specialization.
To address this, EESD distills predictions from a dense EMA ensemble teacher, motivated by~\cite{mun2018learning}, to provide stable supervision for the sparse MoE model, particularly for ambiguous tokens.

When evaluated on CLIP ViT-B/32 and ViT-B/16 models, Cluster-aware Upcycling achieves consistent improvements over existing upcycling methods on several zero-shot and few-shot benchmarks.
More importantly, our analysis quantitatively confirms that the proposed method successfully resolves the problems of expert symmetry and redundancy. 
This structural improvement---characterized by significantly lower inter-expert similarity and more disentangled subspaces---translates directly into enhanced zero-shot and few-shot generalization, thereby demonstrating the clear benefits of structured, semantic-aware initialization.

In summary, our key contributions are organized as:
\begin{itemize}
    \item We propose Cluster-aware Upcycling, a novel initialization strategy that considers latent semantic structures for both expert and router parameters, effectively breaking expert symmetry from the onset of training.
    \item We introduce the EESD loss, which provides stable ensemble-level supervision for tokens with high routing uncertainty, thereby preserving and enhancing expert specialization and robustness.
    \item We demonstrate that Cluster-aware Upcycling consistently outperforms upcycled CLIP models across various zero-shot and few-shot benchmarks, while promoting diverse and disentangled expert representations.
\end{itemize}
\vspace{-1mm}
\section{Background}
\label{sec:background}

\subsection{Mixture-of-Experts} \label{sec:background-moe}

Scaling model size is a well-established path to improving performance.
However, in dense architectures where all parameters are activated for every token, training and inference costs grow proportionally to model size.

Mixture-of-Experts (MoE) architectures offer a sparse alternative that decouples model capacity from the computing cost, thus improving efficiency.
The key idea is to scale up the total number of parameters while activating only a small subset for each input token, such that different tokens follow different computational paths through the model.
This sparsity allows the model to achieve the benefits of massive capacity without a proportional increase in computation, especially at inference time.

In standard dense Transformers, each block contains a feed-forward network (FFN), which is defined as
\begin{equation}
    \operatorname{FFN}(\bm{x}) = f(\bm{x}; \bm{W}),
\end{equation}
where $\bm{W}$ is a learnable parameter matrix.

In MoE architectures, this FFN is replaced by a sparse MoE layer containing $N_e$ expert networks
$\{E_i\}_{i=1}^{N_e}$
and a router.
Other modules, such as attention, are shared across tokens.
Each expert shares the same architecture as the dense FFN but has its own set of parameters $\bm{W}_i$ as:
\begin{equation}
    E_i(\bm{x}) = f(\bm{x}; \bm{W}_i).
\end{equation}
A router, parameterized by $\bm{W}_r$, produces routing probabilities that dispatch each token across experts:
\begin{equation}\label{eq-moe-softmax}
    g(\bm{x}) = \operatorname{softmax}(\bm{W}_r \bm{x}),
\end{equation}
where $g(\bm{x}) \in \mathbb{R}^{N_e}$, and $g_i(\bm{x})$ denotes the routing probability of assigning token $\bm{x}$ to expert $E_i$.
The MoE layer output is given by:
\begin{equation}
    y_{\text{MoE}}(\bm{x}) = \sum_{i \in \mathcal{T}_k(\bm{x})} \tilde g_i(\bm{x})\,E_i(\bm{x}),
\label{eq:moe_output}
\end{equation}
where $\mathcal{T}_k(\bm{x})$ denotes the indices of the top-$k$ entries of $g(\bm{x})$, and $\tilde g_i(\bm{x}) = g_i(\bm{x}) / \sum_{j \in \mathcal{T}_k(\bm{x})} g_j(\bm{x})$ is the renormalized probability over the selected experts.

MoE models are optimized using a task-specific loss $\mathcal{L}_{\text{task}}$, \eg, cross-entropy or contrastive loss, together with an auxiliary load-balancing loss $\mathcal{L}_{\text{lb}}$ that encourages uniform expert utilization and prevents dead experts~\cite{lepikhin2020gshard, fedus2022switch}:
\begin{equation}
    \mathcal{L}_{\text{lb}} = \sum_{i=1}^{N_e} a_i \mathbb{E}_{\bm{x}}[g_i(\bm{x})],
\end{equation}
where $a_i$ denotes the fraction of tokens routed to expert $E_i$.

\begin{figure*}[!t]
    \centering
    \scalebox{0.95}{
        \begin{tabular}{@{}cc@{}}
            \includegraphics[width=1\textwidth]{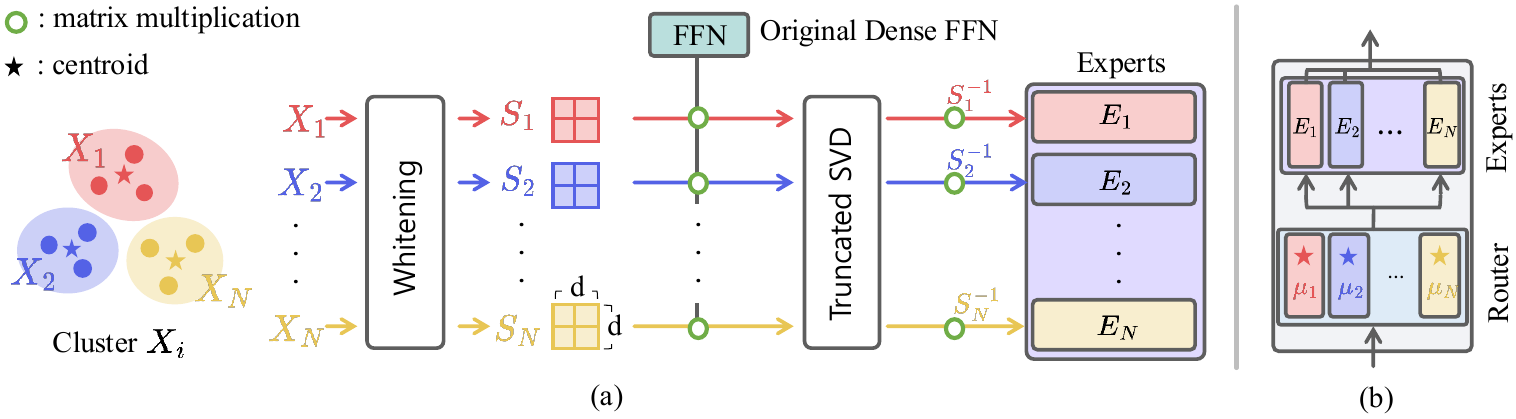}
        \end{tabular}
    }
    \vspace{-1mm}
    \caption{An illustration of how Cluster-aware Upcycling initializes the MoE layer. (a) Input activations are clustered to obtain whitening matrices and cluster centroids, which are used to initialize expert and router parameters, respectively.
    (b) The resulting MoE layer, where experts capture the subspace of their corresponding clusters, and the router aligns with the underlying semantic structure of the data.}
   \vspace{-1mm}
    \label{fig:cluster-upcycling}
\end{figure*}

\subsection{Sparse Upcycling}
Training MoE models from scratch is computationally expensive.
Sparse Upcycling~\cite{komatsuzaki2023sparse} offers a warm-start strategy by transforming a pretrained dense model into an MoE model.
Concretely, each dense FFN is replaced with an MoE layer consisting of $N_e$ experts and a router.
All experts are initialized by copying the parameters from the original dense FFN, \ie, $\forall i, \bm{W}_i \leftarrow \bm{W}$, while the router weight $\bm{W}_r$ is randomly initialized.
All other components of the original dense model remain unchanged.
This replication ensures that the upcycled model is functionally equivalent to the dense model at initialization, providing a warm start for subsequent MoE training.

While Sparse Upcycling efficiently leverages pretrained knowledge, it initializes all experts with identical parameters, which inherently hampers expert diversity.
Since the experts are initially indistinguishable and the router is randomly initialized, establishing a meaningful basis for specialized routing in the early phase is challenging, failing to induce effective expert specialization.
\section{Method}
\label{sec:method}

\begin{figure}[t]
    \centering
    \includegraphics[width=0.8\columnwidth]{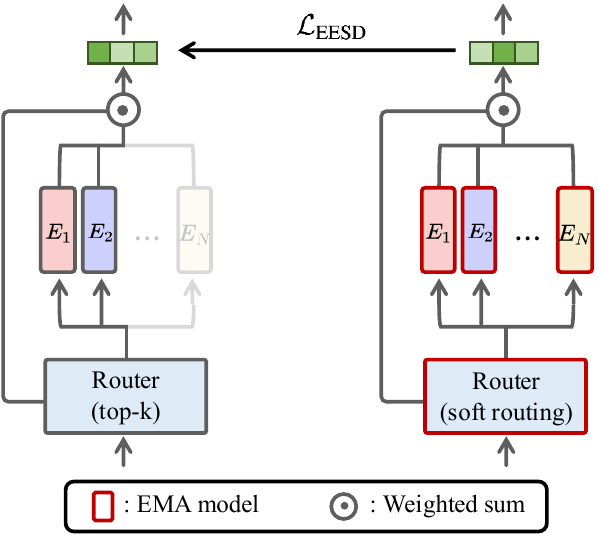}
    \vspace{-1mm}
    \caption{Expert-ensemble self-distillation (EESD).
    The dense EMA ensemble (right) performs soft routing over all experts and provides stable supervision for the sparse top-$k$ MoE model (left), particularly for tokens with high routing uncertainty.}
    \label{fig:eesd}
    \vspace{-1mm}
\end{figure}

Our goal is to achieve expert diversity and structured routing
while preserving the foundational knowledge of the pretrained dense model. 
To this end, we propose Cluster-aware Upcycling, a method that extracts semantic structures from the dense model’s representations to initialize MoE parameters.
The proposed strategy comprises three components: 1) partitioning the dense model's activations into clusters to capture the underlying semantic structure, 
2) initializing expert parameters using the subspace representations of their corresponding clusters,
and 3) initializing the router parameters with the cluster centroids. 
This process alleviates expert symmetry at the onset of training by aligning both experts and the router with the data’s semantic manifold.
Moreover, we introduce the EESD loss, which provides stable ensemble-level supervision particularly for ambiguous tokens.
Figures~\ref{fig:cluster-upcycling} and \ref{fig:eesd} illustrate our cluster-aware initialization and the EESD loss, respectively.

\subsection{Clustering Input Activations}
\label{sec:clustering}
To provide a meaningful basis for expert specialization, we first partition the dense model’s activation space to extract a semantic prior for MoE initialization.
We extract input token activations 
$\bm{X}\!=\!\{\bm{x}_j\}_{j=1}^{M}$,
from each FFN block of the pretrained dense model, using a small calibration dataset that represents the training distribution.
We then perform spherical $k$-means clustering using cosine similarity to group these activations.
This clustering objective is deliberately chosen to align with the router’s logit computation (i.e., $\bm{W}_r \bm{x}$ in Eq.~\eqref{eq-moe-softmax}), which fundamentally measures the directional alignment.
This ensures that activations with similar semantic directions, which should be assigned to the same expert, are clustered together.

Given $\ell_2$-normalized activation vectors, 
the spherical $k$-means clustering objective is defined as
\begin{equation}
    \{\bm{\mu}_i\}_{i=1}^{N_e} = \!\!\!\!\!\!\! \underset{\left\{\hat{\bm{\mu}}_i: \|\hat{\bm{\mu}}_i\|_2 =  1\right\}_{i=1}^{N_e}}{\arg\max} \sum_{j=1}^{M} \max_{i} \hat{\bm{\mu}}_i^T \bm{x}_j,
\end{equation}
where $N_e$ denotes the number of experts (and clusters), and $\bm{\mu}_i$ represents the centroid of the $i^\text{th}$ cluster.
Each activation vector $\bm{x}_j$ is assigned to the cluster $c_i \!=\! \arg\max_i~\bm{\mu}_i^T \bm{x}_j$ that yields the highest cosine similarity.
This process partitions the activation matrix $\bm{X}$ into $N_e$ clusters 
$\{\bm{X}_i\}_{i=1}^{N_e}$ 
with their corresponding centroids 
$\{\bm{\mu}_i\}_{i=1}^{N_e}$, 
which together form a semantic partition of the dense activation manifold, serving as a basis for initializing the experts and the router.

\subsection{Cluster-Aware Expert Initialization}
\label{sec:expert_init}
Given the activation clusters 
$\{\bm{X}_i\}_{i=1}^{N_e}$,
we initialize each expert to specialize in its corresponding cluster.
To formalize this, we jointly optimize expert weights $\left\{\bm{W}_i \right\}_{i=1}^{N_e}$ to minimize the within-cluster reconstruction error between each expert's output and that of the dense FFN, parameterized by $\bm{W}$, while discouraging redundant experts:
\begin{equation}
\label{eq:joint_expert_loss}
    \min_{\{\bm{W}_i\}_{i=1}^{N_e}} 
    \sum_{i=1}^{N_e} 
        \Bigl[ \|\bm{W} \bm{X}_i - \bm{W}_i \bm{X}_i\|_F^2 
        - \gamma\sum_{j \neq i} 
        \|\bm{W} \bm{X}_i - \bm{W}_j \bm{X}_i\|_F^2 \Bigr].
\end{equation}
where $\gamma \!=\! \frac{1}{N_e-1}$.
The first term encourages each expert to approximate the dense model within its cluster. In contrast, the second term discourages experts from collapsing to similar solutions, thereby reducing redundancy and encouraging more specialized expert behaviors.

In practice, rather than directly optimizing Eq.~\eqref{eq:joint_expert_loss} or fine-tuning experts on their clusters, we initialize expert parameters using a data-aware truncated SVD~\cite{wangsvd, wang2025svd}, which preserves principal subspaces associated with each cluster.

Truncated SVD provides a principled low-rank approximation that preserves the leading singular components of a weight matrix.
Given an expert weight $\bm{W}_i$, a standard truncated SVD produces the best rank-$r_i$ approximation $\widetilde{\bm{W}}_i \!= \!T_{r_i}(\operatorname{SVD}(\bm{W}_i))$, where $T_{r_i}(\cdot)$ denotes the truncation function that retains the top-$r_i$ singular directions.
However, this depends solely on the weight matrix and ignores how $\bm{W}_i$ interacts with the input distribution of each cluster.

To address this limitation, data-aware truncated SVD extends the standard formulation by incorporating input statistics.
Specifically, for each cluster $\bm{X}_i$, a whitening matrix $\bm{S}_i$ satisfying $\bm{S}_i \bm{S}_i^T \!=\! \bm{X}_i \bm{X}_i^T$ is obtained by Cholesky decomposition, and truncated SVD is then performed on $\bm{W}_i \bm{S}_i$:
\begin{equation}
    \widetilde{\bm{W}}_i = T_{r_i}(\operatorname{SVD}(\bm{W}_i \bm{S}_i)) \bm{S}_i^{-1}.
\end{equation}
The rank $r_i$ is defined as the effective rank of $\bm{W}_i \bm{S}_i$, where $\sigma_{i,j}$ are its singular values, such that $\sum_{j=1}^{r_i} \!\sigma_{i,j}^2 / \sum_j \!\sigma_{i,j}^2 \!\ge\! \tau$, thereby retaining at least $\tau$ of the total spectral energy.

This formulation ensures that the truncation loss under the data distribution is exactly given by the sum of squared discarded singular values:
\begin{equation}
    \left\|\bm{W}_i \bm{X}_i - \widetilde{\bm{W}}_i \bm{X}_i\right\|_F^2 = \sum_{j>r_i} \sigma_{i,j}^2.
\end{equation}
This implies that the retained components correspond to the principal directions associated with each cluster, while the discarded components correspond to low-energy directions.

\subsection{Cluster-Aware Router Initialization}
\label{sec:router_init}
Since each expert is associated with a cluster, the corresponding cluster centroid provides a natural prior for routing.
Thus, we initialize the router parameters $\bm{W}_r \in \mathbb{R}^{N_e \times d}$ using the $\ell_2$-normalized cluster centroids 
$\{\bm{\mu}_i\}_{i=1}^{N_e}$
obtained from spherical $k$-means clustering:
\begin{equation}
    \bm{W}_r = [\bm{\mu}_1^T; \bm{\mu}_2^T; \ldots; \bm{\mu}_{N_e}^T].
\end{equation}
This initialization aligns early routing decisions with the underlying semantic structure of the data, allowing experts to receive semantically coherent tokens from the start rather than arbitrary assignments induced by random routing.

\subsection{Expert-Ensemble Self-Distillation}
\label{sec:eesd}
Routing probabilities reflect how confidently the router assigns each token to experts.
Tokens with near-uniform routing probabilities, \ie, high routing uncertainty, indicate weak alignment between input and experts, making it difficult for model to reinforce consistent expert specialization.

To preserve and reinforce the experts' specialization seeded by cluster-aware initialization, we introduce an expert-ensemble self-distillation (EESD) loss, which is motivated by~\cite{mun2018learning}.
This loss uses a dense EMA ensemble to provide stable supervision for the sparse MoE model, particularly when routing decisions are uncertain.

Specifically, we construct a teacher model by applying exponential moving average (EMA) updates to the MoE parameters.
Unlike the sparse activations in a standard MoE layer, this teacher operates in a dense, full-capacity mode by activating all experts simultaneously:
\begin{equation}
    y_{\text{ens}}(\bm{x}) = \sum_{i=1}^{N_e} g_i^{\text{ema}}(\bm{x})\,E_i^{\text{ema}}(\bm{x}),
\end{equation}
where $g_i^{\text{ema}}$ and $E_i^{\text{ema}}$ denote EMA router and expert, respectively, whose parameters are updated as $\bm{W_i}^{\text{ema}} \leftarrow \beta \bm{W_i}^{\text{ema}} + (1-\beta)\bm{W_i}$.
Since this teacher aggregates the knowledge of all EMA experts, it provides a stable distillation target that the top-$k$ prediction aims to approximate.
Recall that, in contrast, the MoE layer prediction is computed using only the top-$k$ activated experts as defined in Eq.~\eqref{eq:moe_output}.

%

\begin{table*}[!t]
    \centering
    \renewcommand{\arraystretch}{1.05}
    \setlength{\tabcolsep}{6.5pt}
    \caption{Comparison of upcycling methods on CLIP ViT-B/32 and ViT-B/16 across zero-shot retrieval and classification benchmarks, measured by Recall@1 and Accuracy, respectively. Cluster-aware Upcycling achieves the best performance across most benchmarks.}
    \vspace{-1mm}
    \label{table:zeroshot}
    \scalebox{0.85}{
    \begin{tabular}{lclccccccccccc}
    \toprule
    \multicolumn{3}{c}{Model} & 
    \multicolumn{3}{c}{MSCOCO} & 
    \multicolumn{7}{c}{ImageNet-1k} &
    \multicolumn{1}{c}{VTAB} \\
    \cmidrule(lr){1-3} \cmidrule(lr){4-6} \cmidrule(lr){7-13} \cmidrule(lr){14-14}
    Arch. & MoE Init & Samples & I$\rightarrow$T & T$\rightarrow$I & Avg. & Val & V2 & A & R & Sketch & ObjNet & Avg. & Natural \\

    \midrule
    \rowcolor{gray!15} \multicolumn{14}{c}{ViT-B/32} \\
    
    \multirow{2}{*}{Dense} & \multirow{2}{*}{-} & 4.0B & 25.5 & 42.5 & 34.0 & 49.6 & 41.9 & 9.7 & 56.7 & 34.9 & 31.0 & 37.3 & 52.4 \\
    
    & & 4.0B+1.3B & 30.8 & 47.5 & 39.2 & 56.7 & 48.5 & 13.9 & 64.0 & 41.2 & 36.1 & 43.4 & 58.3 \\
    \midrule

    \multirow{5}{*}{MoE}  & Drop-Upcycling~\cite{nakamura2025dropupcycling} & \multirow{5}{*}{4.0B+1.3B} & 29.7 & 46.5 & 38.1 & 56.0 & 47.7 & 12.9 & 63.4 & 40.8 & 34.3 & 42.5 & 57.8 \\

    & Sparse Upcycling~\cite{komatsuzaki2023sparse} & & 30.8 & {48.0} & {39.4} & 57.1 & {49.1} & 13.8 & 64.3 & 41.8 & 36.0 & 43.7 & 58.0 \\

    & CLIP-MoE~\cite{zhang2025clip} & & 29.5 & 46.8 & 38.2 & 56.6 & 48.1 & \textbf{14.3} & 64.2 & 41.4 & 35.7 & 43.4 & {58.8} \\
    
    & DeRS-LM~\cite{huang2025ders} & & \textbf{31.0} & 47.7 & {39.4} & 56.8 & 48.6 & 13.9 & 64.2 & 41.1 & {36.4} & 43.5 & 58.1 \\
    
    & \textbf{Cluster-aware Upcycling} & & \textbf{31.0} & \textbf{48.2} &\textbf{39.6} & \textbf{57.3} & \textbf{49.2} & 14.0 & \textbf{65.2} & \textbf{42.3} & \textbf{36.5} & \textbf{44.1} & \textbf{59.1} \\
    
    \midrule
    \midrule

    \rowcolor{gray!15} \multicolumn{14}{c}{ViT-B/16} \\
    \multirow{2}{*}{Dense} & \multirow{2}{*}{-} & 4.0B & 32.5 & 49.1 & 40.8 & 59.4 & 51.7 & 20.1 & 67.3 & 42.9 & 39.4 & 46.8 & 58.1 \\
    
    & & 4.0B+1.3B & 34.3 & 50.8 & 42.6 & 62.5 & 54.4 & 23.5 & 70.6 & 45.8 & 42.5 & 49.9 & 62.6  \\
    \midrule
    
    \multirow{4}{*}{MoE} & Drop-Upcycling~\cite{nakamura2025dropupcycling} & \multirow{4}{*}{4.0B+1.3B} & 34.1 & 51.3 & 42.7 & 62.0 & 54.5 & 22.7 & 70.8 & 45.7 & 42.9 & 49.8 & 60.9 \\

    & Sparse Upcycling~\cite{komatsuzaki2023sparse} & & {34.9} & 50.9 & 42.9 & {63.0} & \textbf{55.1} & 23.7 & 71.2 & 46.3 & 42.3 & 50.3 & 62.0 \\

    & CLIP-MoE~\cite{zhang2025clip} & & 34.0 & 51.5 & 42.8 & 62.9 & 54.9 & \textbf{24.5} & {71.6} & 46.2 & {43.4} & {50.6} & {62.8} \\

    & \textbf{Cluster-aware Upcycling} & & \textbf{35.4} & \textbf{51.6} & \textbf{43.5} & \textbf{63.2} & \textbf{55.1} & {24.1} & \textbf{72.1} & \textbf{46.8} & \textbf{43.5} & \textbf{50.8} & \textbf{63.3}  \\
    
    \bottomrule
    \vspace{-1mm}
    \end{tabular}}
\end{table*}
%

The EESD loss minimizes the discrepancy between the outputs of the sparse MoE and dense EMA ensemble as:
\begin{equation}
    \mathcal{L}_{\text{EESD}}
    = \frac{1}{T} \sum_{\bm{x}} \big\|\,\operatorname{sg}(y_{\text{ens}}(\bm{x})) - y_{\text{MoE}}(\bm{x}) \big\|_2^2,
\end{equation}
where $T$ is the total number of tokens, and $\operatorname{sg}(\cdot)$ denotes the stop-gradient operator, which prevents gradients from flowing into the teacher prediction.

When routing probabilities are nearly uniform, the discrepancy between the mixture of top-$k$ output and the dense ensemble prediction tends to be larger, and the loss provides stronger guidance.
Conversely, for confident tokens with sharp routing probabilities, the top-$k$ output closely aligns with the ensemble prediction, so the loss remains small and does not interfere with expert specialization.

The overall training objective combines the task, load-balancing, and EESD losses:
\begin{equation}
    \mathcal{L}
    = \mathcal{L}_{\text{task}}
     + \lambda_{\text{lb}}\mathcal{L}_{\text{lb}}
     + \lambda_{\text{EESD}}\mathcal{L}_{\text{EESD}}.
\end{equation}
\section{Experiments}

\subsection{Implementation Details}\label{sec:imp}
We use CLIP~\cite{radford2021learning} ViT-B/16 and ViT-B/32 as our dense baselines.
For dense pretraining, we follow the public CLIP configuration trained on LAION-400M~\cite{schuhmann2021laion}, with patch dropout 0.5 and 5.3B seen samples over 20 epochs.
The global batch size is set to 16$K$ for ViT-B/32 and 65$K$ for ViT-B/16, with linear warmup for the first 2\% of total steps followed by linear decay.
The maximum learning rate is 0.005, and the minimum is 0.

For MoE upcycling, we use the dense model parameters from the checkpoint at 15 epochs, corresponding to 4B seen samples, and replace every other FFN with an MoE layer.
We use the DeepSpeed-MoE~\cite{rajbhandari2022deepspeed} implementation with token-choice, top-2 routing, and 8 experts per layer.
The capacity factor is set to 1.5 for MoE layers in the B/16 model and 2.0 in the B/32 model during training, and 2.0 for both at inference.
Each upcycled model is trained on LAION-400M for 1.3B seen samples over 5 epochs, using global batch sizes of 16$K$ for B/32 and 32$K$ for B/16.
The learning rate schedule uses 2\% linear warmup followed by linear decay to zero, with the peak learning rate set to match the dense model's learning rate at the upcycling checkpoint, following~\cite{komatsuzaki2023sparse}.
Optimizer states are reinitialized rather than loaded from the dense model checkpoint.
The load-balancing loss coefficient $\lambda_{\text{lb}}$ is set to 0.001.

For clustering, we reduce the activation dimensionality by a factor of eight using PCA, and then perform spherical k-means clustering using Faiss~\cite{johnson2019billion} on 128$K$ sampled image–text pairs from training set.
For expert initialization, we set the effective rank $r_i$ to the smallest value that preserves at least a $\tau\!=\!0.95$ fraction of the spectral energy, while ensuring that $r_i$ is larger than half of the full rank.
Cluster-aware expert initialization is applied only to the first linear layer of each expert FFN, as clustering is performed on the FFN input activations and the identified cluster structure is not directly aligned in subsequent layers.
For Expert-Ensemble Self-Distillation, we use an EMA coefficient $\beta$ of 0.999 and set $\lambda_{\text{EESD}}$ to 1.0 for ViT-B/32 and 0.1 for ViT-B/16, excluding padding tokens in the loss computation.

\subsection{Evaluation Setup}
We evaluate our method using CLIP-Benchmark~\cite{cherti_2025_15403103}, which covers zero-shot retrieval, zero-shot classification, few-shot classification, and full fine-tuning tasks.

We compare our method with several upcycling baselines.
Sparse Upcycling~\cite{komatsuzaki2023sparse} copies the dense FFN weights to all experts while initializing the router randomly.
Drop-Upcycling~\cite{nakamura2025dropupcycling} partially reinitializes expert channels to introduce diversity with a randomly initialized router.
DeRS-LM~\cite{huang2025ders} employs an expert-shared base weight and represents experts as low-rank matrices.
CLIP-MoE~\cite{zhang2025clip} introduces a multi-stage contrastive learning strategy tailored for MoE upcycling in CLIP.
All methods use the same MoE configuration, dataset, and training schedule as in Section~\ref{sec:imp} to ensure a fair comparison.
Experiments are conducted on 64 H200 GPUs.

\subsection{Quantitative Results}
%

\begin{table}[!t]
    \centering
    \renewcommand{\arraystretch}{1.05}
    \setlength{\tabcolsep}{7pt}
    \vspace{-1mm}
    \caption{ImageNet-1k few-shot and full fine-tuning results for the upcycled ViT-B/16 model. 
    Cluster-aware Upcycling consistently outperforms other upcycling methods.}
    \label{table:finetuning}
    \scalebox{0.85}{
    \begin{tabular}{lcccc}
    \toprule
    \multicolumn{2}{c}{Model} & 
    \multicolumn{3}{c}{ImageNet-1k} \\
    \cmidrule(lr){1-2} \cmidrule(lr){3-5}
    Arch. & MoE Init & 5-shot & 10-shot & FT \\
    \midrule
    Dense & - & 50.4 & 57.1 & 72.8 \\
    \midrule
    \multirow{4}{*}{MoE} & Sparse Upcycling~\cite{komatsuzaki2023sparse} & 50.9 & 57.8 & 73.0 \\
    
    & Drop-Upcycling~\cite{nakamura2025dropupcycling} & 51.1& 57.9 & 73.1 \\

    & CLIP-MoE~\cite{zhang2025clip} & 51.3 & 58.0& 73.2 \\
    
    & \textbf{Cluster-aware Upcycling} & \textbf{51.5} & \textbf{58.2} & \textbf{73.3} \\
    \bottomrule
    \end{tabular}}
    \vspace{-1mm}
\end{table}
\subsubsection{Zero-shot Results}

Table~\ref{table:zeroshot} summarizes zero-shot cross-modal retrieval and classification performance for the upcycled ViT-B/32 and ViT-B/16 models.

For the upcycled ViT-B/32 model, Cluster-aware Upcycling achieves the strongest overall performance, while the baselines do not exhibit consistent gains across benchmarks.
In Drop-Upcycling, partial reinitialization perturbs the pretrained representation more aggressively, which appears to weaken the benefits of warm-start upcycling and leads to lower overall performance.
Sparse Upcycling generally provides competitive results but shows noticeably lower performance on VTAB-Natural, suggesting weaker generalization to out-of-distribution.
CLIP-MoE, which is tailored for contrastive learning, achieves strong results on several benchmarks, \eg, ImageNet-A and VTAB-Natural, but its cross-modal retrieval performance falls even below the dense baseline.
DeRS-LM achieves competitive results on some benchmarks despite using low-rank experts, but performs worse on average.
In contrast, Cluster-aware Upcycling achieves the best performance on most benchmarks.

A similar trend is observed for the ViT-B/16, where the performance gap becomes even more pronounced.
These improvements arise from the combined effect of cluster-aware initialization together with EESD, which collectively encourage early expert specialization and maintain it throughout training.

This specialization is consistent with the improved out-of-distribution generalization observed in VTAB-Natural.
Moreover, the advantage becomes clearer for larger models and as training progresses (see Supplementary Section~\ref{sec:supple_training_dynamics}), further supporting the scalability of our approach.

%

\begin{table}[t]
    \centering
    \renewcommand{\arraystretch}{1.05}
    \setlength{\tabcolsep}{9.5pt}
    \caption{Ablation study on cluster-aware initialization and EESD loss for the upcycled ViT-B/16 model.}
    \vspace{-1mm}
    \label{table:ablation}
    \scalebox{0.85}{
    \begin{tabular}{cccccc}
        \toprule
        \multicolumn{2}{c}{Model} & 
        \multicolumn{2}{c}{MSCOCO} & 
        \multicolumn{2}{c}{ImageNet-1k} \\
        
        \cmidrule(lr){1-2} \cmidrule(lr){3-4} \cmidrule(lr){5-6}
        Cluster-init. & EESD & I$\rightarrow$T & T$\rightarrow$I & Val & 10-shot \\
        \midrule
         &  &  34.9 & 50.9 & 63.0 & 57.8 \\
        \checkmark & & 35.1 & 51.1 & \textbf{63.2} & 58.1 \\
         & \checkmark & 34.6 & 51.4 & 62.7 & 57.8 \\
        \checkmark & \checkmark & \textbf{35.4} & \textbf{51.6} & \textbf{63.2} & \textbf{58.2} \\
        \bottomrule
    \end{tabular}}
    \vspace{-1mm}
\end{table}
\subsubsection{Fine-tuning Results}
Table~\ref{table:finetuning} presents 5-shot, 10-shot, and full fine-tuning accuracy on ImageNet for the upcycled ViT-B/16. 
Cluster-aware Upcycling consistently outperforms comparison methods across all settings. 
The improvements are most pronounced in few-shot regimes, where initialization quality is critical due to limited training signals. 
As more labeled data becomes available, the relative advantage becomes smaller but remains consistent, indicating that cluster-aware initialization provides a stronger starting point that persists throughout adaptation.
These results suggest that semantic structure at initialization yields expert representations that generalize better not only in zero-shot transfer but also across diverse fine-tuning settings.
%
\begin{figure*}[!t]
    \centering
    \begin{subfigure}[t]{0.24\linewidth}
        \centering
        \includegraphics[width=\linewidth]{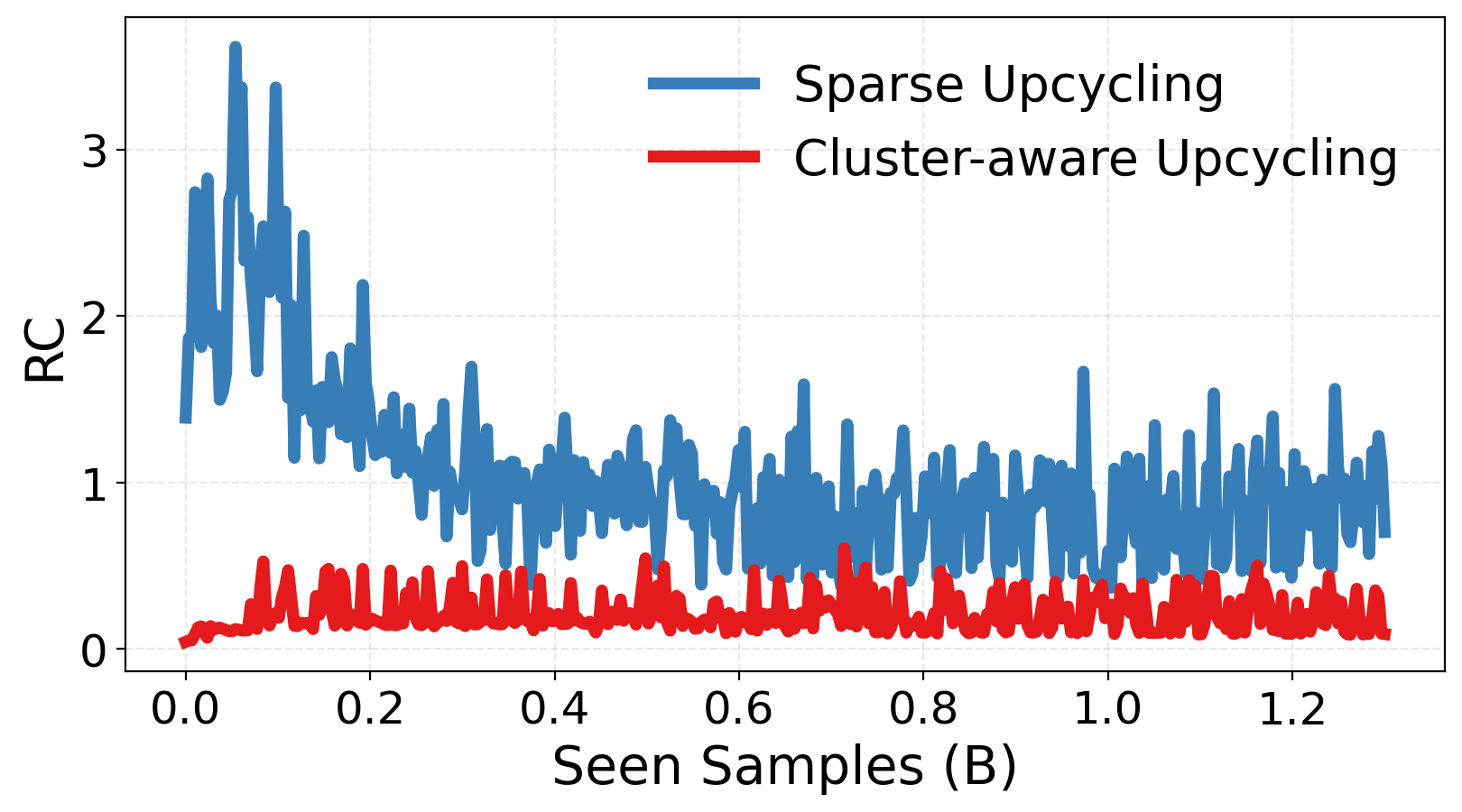}
        \caption{Relative Compactness}
        \label{fig:rc}
    \end{subfigure}
    \hfill
    \begin{subfigure}[t]{0.24\linewidth}
        \centering
        \includegraphics[width=\linewidth]{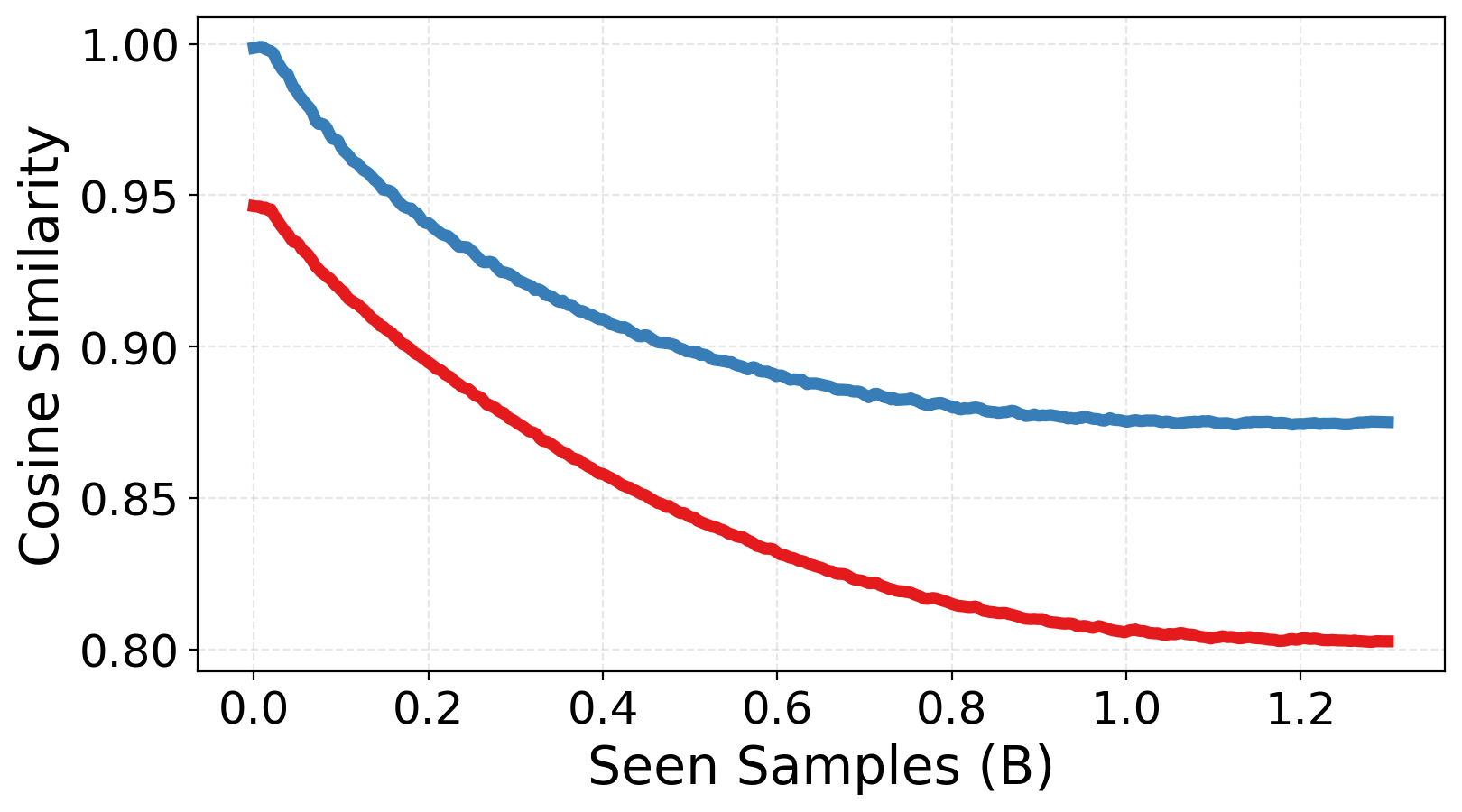}
        \caption{Expert Similarity}
        \label{fig:expert_similarity}
    \end{subfigure}
    \hfill
    \begin{subfigure}[t]{0.24\linewidth}
        \centering
        \includegraphics[width=\linewidth]{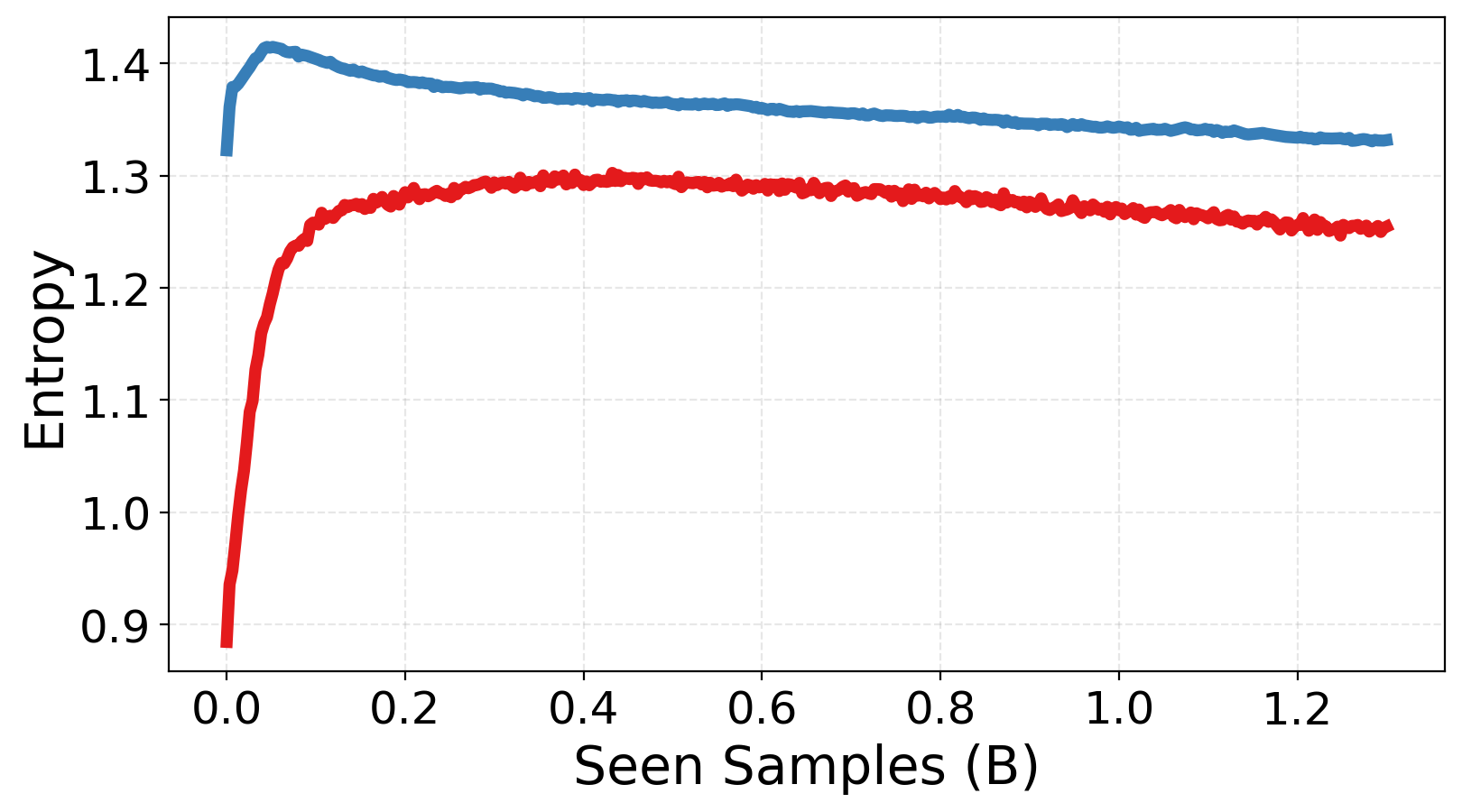}
        \caption{Routing Entropy}
        \label{fig:analysis_routing_entropy}
    \end{subfigure}
    \hfill
    \begin{subfigure}[t]{0.24\linewidth}
        \centering
        \includegraphics[width=\linewidth]{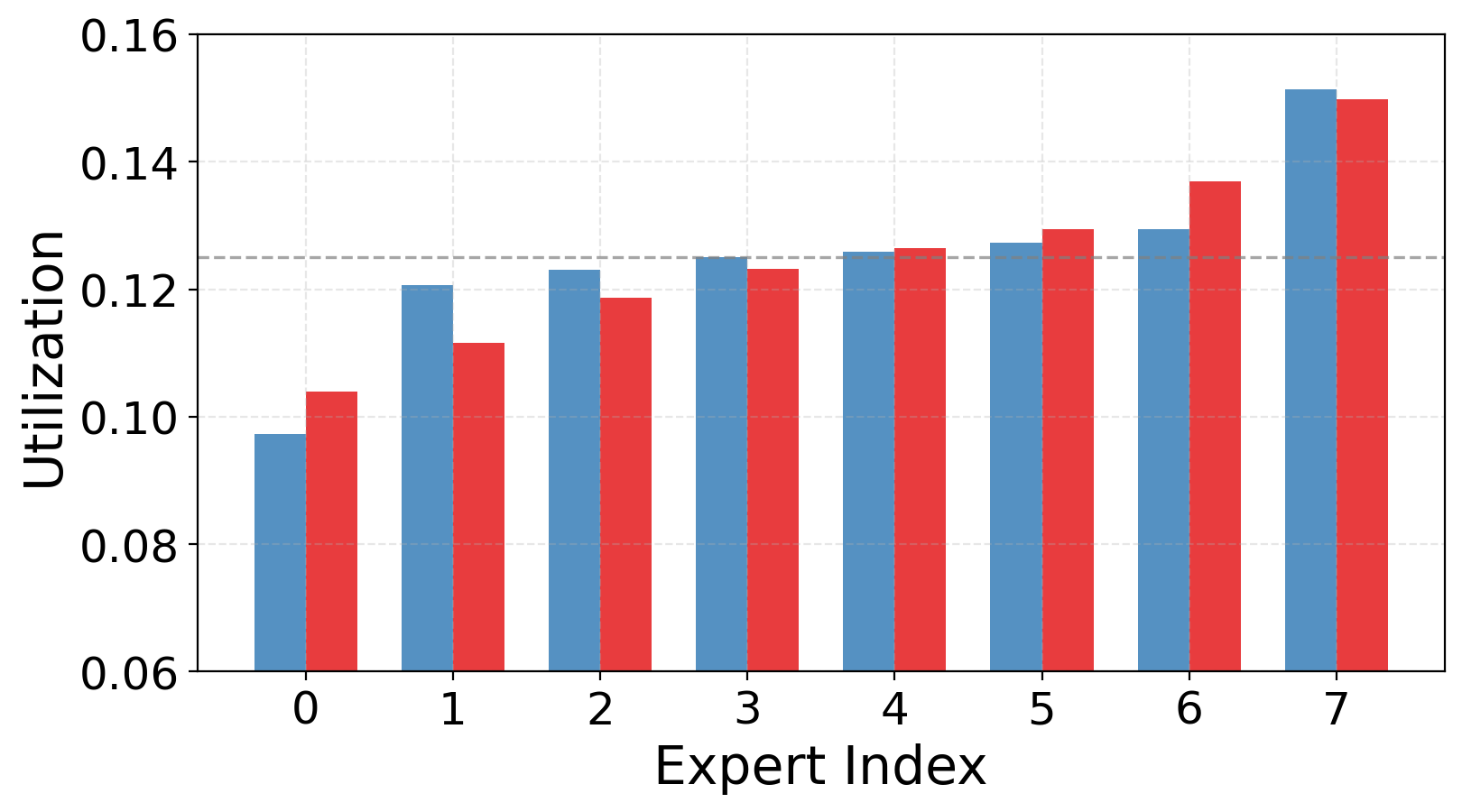}
        \caption{Expert utilization}
        \label{fig:analysis_expert_utilization}
    \end{subfigure}
    \vspace{-1mm}
    \caption{
        (a) Relative Compactness measures the overlap between intra- and inter-expert variance, where lower values indicate more disentangled subspaces. 
        (b) Expert similarity shows pairwise cosine similarity between expert weights, with Cluster-aware Upcycling maintaining higher parameter diversity. 
        (c) Routing entropy captures routing uncertainty, with our model achieving lower entropy and more stable expert assignments.
        (d) Expert utilization patterns across experts, showing balanced routing without routing collapse.
        }
    \label{figure:analysis}
    \vspace{-1mm}
\end{figure*}

\subsection{Ablation Study}
We evaluate the contribution of the two proposed components, cluster-aware initialization and the EESD loss. 
As shown in Table~\ref{table:ablation}, cluster-aware initialization alone provides consistent improvements over the Sparse Upcycling baseline across both retrieval and classification metrics. 
In contrast, the standalone effect of EESD is modest and does not show a clear or consistent improvement trend.
However, when combined with cluster-aware initialization, EESD yields further gains across all benchmarks, indicating that the two components play complementary roles.
More specifically, cluster-aware initialization establishes a meaningful basis for early specialization by breaking expert symmetry, while EESD helps preserve and reinforce this specialization during training by providing ensemble-level guidance under uncertain routing.

Because the EMA teacher aggregates only the expert outputs instead of running a full-model ensemble, the EESD loss introduces only modest overhead, approximately 5.3\% in wall-clock time and 2.8\% in memory in our experiments.
This design makes EESD practical even for large-scale MoE architectures, where full model ensembles would otherwise be prohibitive.

\subsection{Analysis}
\label{sec:analysis}
To understand how Cluster-aware Upcycling influences expert specialization, we analyze four aspects of the trained MoE model, relative compactness, expert diversity, routing entropy, and expert utilization, summarized in Figure~\ref{figure:analysis}.

\paragraph{Relative compactness}
To assess how experts structure their feature subspaces during training, we measure the Relative Compactness (RC).
Specifically, we compute the within-expert covariance $\Sigma_{\bm{W}}$ as the covariance of token representations within each expert, and the between-expert covariance $\Sigma_{\bm{B}}$ as the covariance of the mean output vectors of all experts.
RC is then computed as $\mathrm{RC} \!= \!\operatorname{Tr}(\Sigma_{\bm{W}} \Sigma_{\bm{B}}^{\dagger})$, which measures how strongly the within-expert variance aligns with the between-expert variance directions.
Lower RC values indicate that each expert’s internal variance lies in directions orthogonal to those of other experts, implying disentangled and non-redundant expert subspaces. 
As shown in Figure~\ref{fig:rc}, our model consistently yields lower RC throughout training, indicating geometrically independent expert representations rather than overlapping or redundant ones.

\paragraph{Expert diversity}
We assess expert diversity through pairwise cosine similarity between expert parameters~\cite{lo2025closer}. 
As illustrated in Figure~\ref{fig:expert_similarity}, Sparse Upcycling exhibits high similarity among experts, indicating limited diversification. 
In contrast, Cluster-aware Upcycling shows lower similarity, indicating that the experts are more clearly separated in weight space, consistent with the disentangled latent subspaces captured by relative compactness analysis.

\paragraph{Routing entropy}
\label{sec:analysis_routing_entropy}
We analyze routing entropy to assess how confidently the model assigns tokens to experts. 
As shown in Figure~\ref{fig:analysis_routing_entropy}, our model begins with low entropy due to the cluster-aware initialization, increases gradually during training as the load-balancing loss encourages exploration, and eventually stabilizes at a lower level.
In contrast, the model trained with Sparse Upcycling maintains consistently higher entropy throughout training, indicating less confident and less specialized routing behavior.

\paragraph{Expert utilization}
\label{sec:analysis_expert_utiliaztion}
%

\begin{figure}[!t]
    \centering
    \begin{subfigure}[t]{0.48\linewidth}
        \centering
        \includegraphics[width=\linewidth]{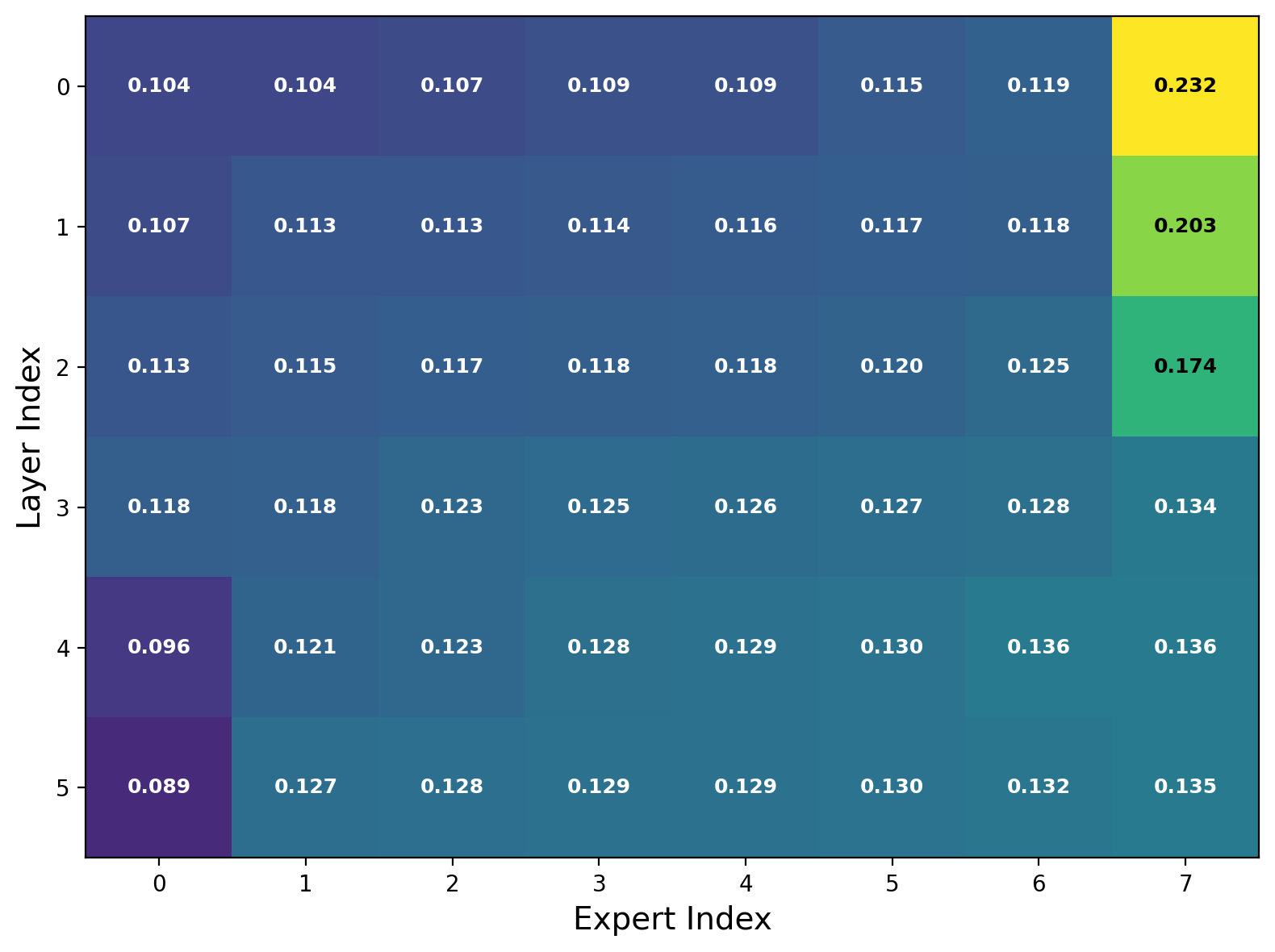}
        \caption{Sparse Upcycling}
    \end{subfigure}
    \hfill
    \begin{subfigure}[t]{0.48\linewidth}
        \centering
        \includegraphics[width=\linewidth]{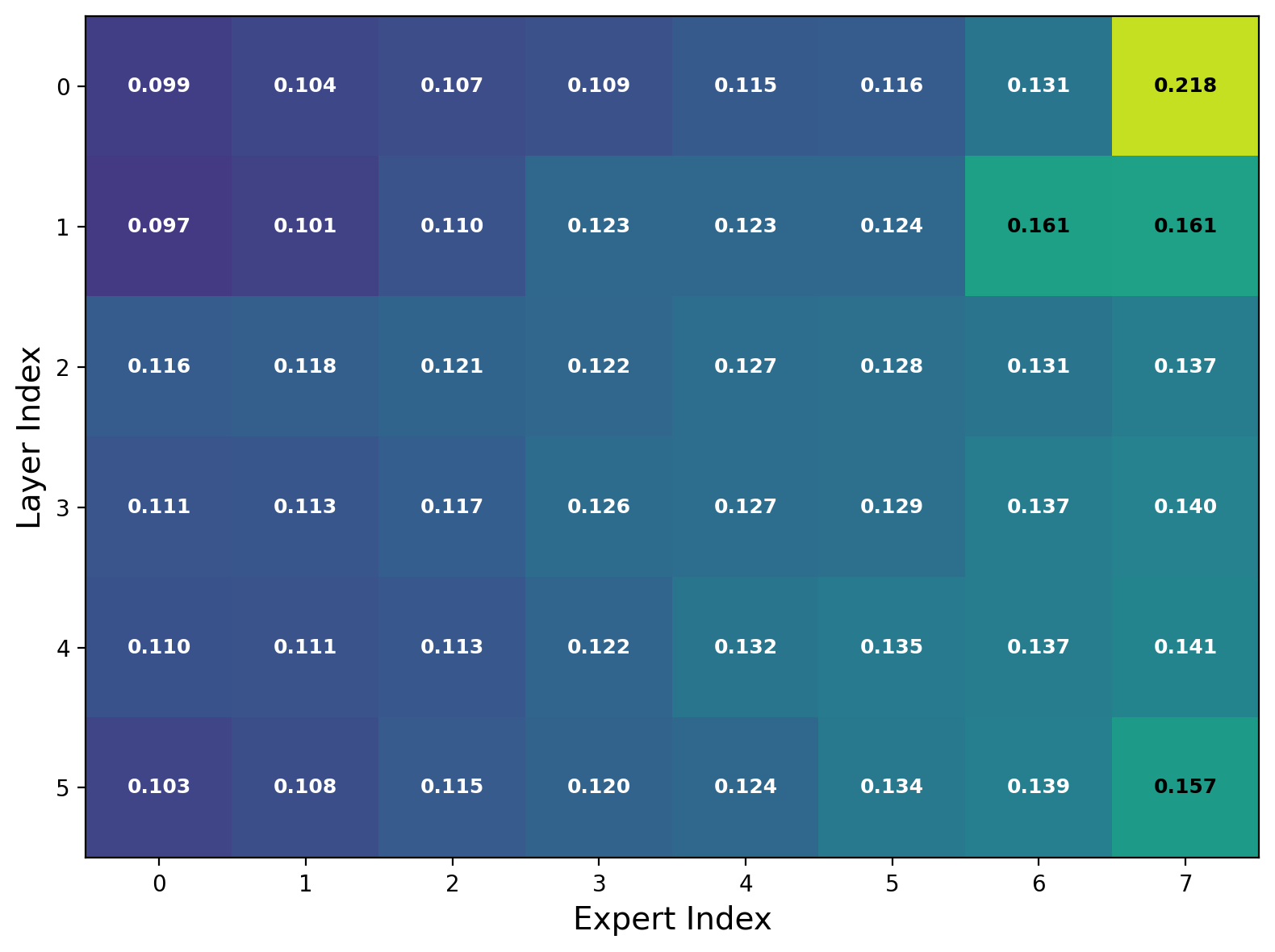}
        \caption{Cluster-aware Upcycling}
    \end{subfigure}
    \vspace{-1mm}
    \caption{Detailed expert utilization across mixture-of-experts layers in vision encoders. Best viewed in color.}
    \label{figure:analysis_routingpattern_vision}
    \vspace{-1mm}
\end{figure}
We analyze expert utilization across layers by measuring the fraction of tokens assigned to each expert.
As shown in Figure~\ref{fig:analysis_expert_utilization}, both Sparse Upcycling and Cluster-aware Upcycling maintain balanced utilization across experts, indicating that no routing collapse occurs.
Furthermore, our method exhibits more stable utilization patterns across layers, suggesting that the proposed method enables structured specialization without inducing routing imbalance. 
The detailed layer-wise patterns provided in Figure~\ref{figure:analysis_routingpattern_vision} further show that Cluster-aware Upcycling maintains slightly more diverse yet still well-balanced expert utilization patterns across layers.

\section{Related Work}
\paragraph{Mixture-of-Experts}
Mixture-of-Experts (MoE) architectures enable efficient scaling by activating only a subset of expert networks for each input, rather than processing through all parameters. 
This conditional computation strategy has proven effective across language models~\cite{lepikhin2020gshard, fedus2022switch, zoph2022st, jiang2024mixtral}, vision~\cite{riquelme2021scaling, wang2025clip}, and multimodal architectures~\cite{lin2024moe, bai2025qwen2, team2025kimi}. 
In a typical MoE layer, a gating network routes each token to a small number of experts that process the input in parallel, while leaving other experts inactive.
This approach allows models to scale total parameter count while keeping computational cost per token approximately constant.
Recent architectural improvements refine this framework by combining shared and specialized experts~\cite{rajbhandari2022deepspeed, dai2024deepseekmoe}, introducing hierarchical expert organizations~\cite{rajbhandari2022deepspeed}, or replacing experts with more efficient representations such as lookup-table-based structures~\cite{jiemixture}. 

Despite these advances, MoE systems still face challenges in routing stability and expert specialization. Load-balancing losses promote expert usage but may reduce diversity, motivating techniques such as orthogonality constraints, variance-based regularization~\cite{guo2025advancing}, and auxiliary-loss–free routing strategies~\cite{wang2024auxiliary}. 
MoE models also exhibit redundancy among experts, especially when initialized from dense checkpoints, leading to overlapping expert functions. 
Decomposition-based methods~\cite{gudelta, limoe, huang2025ders, fanmake, yuan2025moore} demonstrate that experts can be approximated using shared bases with small residuals, revealing that expert specialization is often weaker than desired.

\paragraph{Mixture-of-Experts Upcycling}
Sparse Upcycling~\cite{komatsuzaki2023sparse} initializes a sparse MoE model by reusing weights from a pre-trained dense model, rather than training from scratch, substantially reducing training cost.
However, since all experts start from identical weights, Sparse Upcycling often results in limited expert diversity and poor specialization.
Empirical analysis confirms this tendency, showing that higher expert similarity degrades performance, whereas models trained from scratch maintain low similarity due to random initialization~\cite{wei2024skywork, lo2025closer}.
To mitigate this issue, subsequent works have explored various strategies to enhance expert diversity, including injecting Gaussian noise into router or expert weights~\cite{komatsuzaki2023sparse, muennighoff2025olmoe, liew2025scaling}, permuting or partially reinitializing feed-forward networks~\cite{komatsuzaki2023sparse, chi2022representation, team2024qwen2}, and adjusting learning rates across components~\cite{komatsuzaki2023sparse}; however, these approaches fail to bring noticeable improvement in expert specialization.
BTX~\cite{sukhbaatarbranch} and Nexus~\cite{gritsch2024nexus} instead fine-tune pre-trained dense language models on multiple domains and use them to initialize expert parameters, introducing domain-level specialization among experts.
Several works also apply specialized distillation to guide expert divergence after initialization during reinforcement learning stages~\cite{huang2026step, xiao2026mimo, liu2025deepseek, zeng2025glm}.
Although effective, these methods rely on explicit domain partitioning and several fine-tuning stages to achieve specialization.
DeRS-LM~\cite{huang2025ders} employs an expert-shared base weight and represents experts as low-rank matrices; however, it does not address expert symmetry during upcycling.
Drop-Upcycling~\cite{nakamura2025dropupcycling} initializes a subset of expert weights with dense weights, while the remaining subset is re-initialized by sampling from a Gaussian distribution parameterized by the original parameters’ estimated statistics, partially alleviating the redundancy issue.
For vision-language models, CLIP-UP~\cite{wang2025clip} applied Sparse Upcycling to CLIP. 
CLIP-MoE~\cite{zhang2025clip} introduces multi-stage contrastive learning for MoE upcycling, though it is limited to a contrastive learning objective.

\paragraph{Clustering perspective on Mixture-of-Experts}
MoE architectures can be viewed through the lens of differentiable clustering, where the router assigns tokens to experts according to similarity in representation space. 
From this viewpoint, experts behave as learnable cluster centroids, and routing defines a soft partition of the token distribution.
Earlier work provides theoretical support for this interpretation.
\citet{chen2022towards} shows that gradient descent on nonlinear MoEs can recover latent cluster structure.
\citet{chi2022representation} demonstrates that router logits form manifolds around expert embeddings.
\citet{dikkala2023benefits} show that routers initialized from random training samples can recover distinct clusters in well-separated mixtures.
Building on this line of thought, ACMoE~\cite{nielsentight} extends this perspective by introducing an adaptive clustering router that enhances expert specialization by scaling features based on cluster tightness. 
Collectively, these studies suggest that the geometry of the representation space strongly shapes MoE behavior and that cluster-aware initialization can promote better specialization and training stability, motivating our method.
\section{Conclusion}

In this work, we introduced Cluster-aware Upcycling, a method that mitigates the expert symmetry problem in MoE upcycling.
Rather than replicating pretrained dense weights identically across experts, our method partitions the activation space into semantic clusters and uses them to provide a meaningful basis for expert and router initialization.
The data-aware truncated SVD preserves pretrained knowledge within each cluster while promoting diversity across experts, and the cluster-informed router initialization aligns early routing with the underlying representation structure. 
In addition, EESD preserves and enhances expert specialization by providing ensemble-level supervision, especially when routing remains ambiguous.
Experiments on upcycled CLIP models show that Cluster-aware Upcycling consistently improves over baseline upcycling methods on zero-shot and few-shot benchmarks. 
Our analysis also reveals lower inter-expert similarity, more disentangled expert subspaces, and more specialized yet balanced routing dynamics, demonstrating that leveraging the semantic structure of the pretrained dense model offers a principled path to effective expert specialization in MoE upcycling.
These results suggest that incorporating semantic structure into MoE upcycling is a simple yet powerful strategy that may generalizes beyond vision-language pretraining.

\clearpage
\paragraph{Acknowledgements}
This work was supported in part by the National Research Foundation of Korea (NRF) [RS-2022-NR070855, Trustworthy Artificial Intelligence], Institute of Information \& Communications Technology Planning \& Evaluation (IITP) [RS2022-II220959 (No.2022-0-00959), (Part 2) Few-Shot Learning of Causal Inference in Vision and Language for Decision Making; RS-2025-25442338, AI Star Fellowship Support Program (Seoul National Univ.); No.RS-2021-II211343, Artificial Intelligence Graduate School Program (Seoul National University)], funded by the Korea government (MSIT).
It was also partly supported by AI-driven Scientific Safety e-Report Information Analysis Technology Development Program [RS-2024-00509777, AI-driven Safety e-Reporting and Value Assessment Technology] funded by the Ministry of the Interior and Safety (MOIS).

{
    \small
    \bibliographystyle{ieeenat_fullname}
    \bibliography{main}
}

\clearpage
\clearpage
\maketitlesupplementary
\setcounter{page}{1}
\setcounter{section}{0}
\renewcommand{\thesection}{\Alph{section}}
\renewcommand{\thefigure}{\Alph{section}.\arabic{figure}}
\counterwithin{figure}{section}
\renewcommand{\thetable}{\Alph{section}.\arabic{table}}
\counterwithin{table}{section}
\renewcommand{\theequation}{\Alph{section}.\arabic{equation}}
\counterwithin{equation}{section}

\section{Additional Quantitative Results}
To complement the comparisons presented in the main paper, we provide additional quantitative results on the upcycled ViT-B/32 in Table~\ref{table:supplementary_full_comparison_b32}.
We first revisit Drop-Upcycling~\cite{nakamura2025dropupcycling}, which reinitializes a randomly selected fraction $r$ of channels using Gaussian noise estimated from the pretrained statistics. 
While the original work reports results with $r\!\!=\!\!0.5$, we observe that smaller perturbation ratios generally lead to better performance across most benchmarks. 
This suggests that aggressively perturbing pretrained weights can degrade the underlying representation learned during pretraining.
However, even under its best-performing configuration ($r\!=\!0.25$), Drop-Upcycling generally underperforms Sparse Upcycling~\cite{komatsuzaki2023sparse}.

To explore a more structured alternative to random perturbation, we introduce a simple variant termed Drop-SVD.
Instead of randomly perturbing channels, Drop-SVD applies singular value decomposition to each weight matrix and reinitializes only the singular vectors corresponding to the lowest 25\% of the spectrum. 
This preserves the dominant semantic subspace encoded in the pretrained weights while introducing diversity through controlled perturbations in the least informative directions.
In contrast to Drop-Upcycling, Drop-SVD achieves performance comparable to Sparse Upcycling and even surpasses it on several benchmarks.
These results indicate that respecting the structure of the pretrained representations is important when designing initialization strategies for MoE upcycling.

Nevertheless, perturbation-based approaches do not explicitly leverage the semantic structure of the pretrained representation space.
In contrast, Cluster-aware Upcycling directly leverages activation-space clustering to initialize experts according to the structure of the pretrained representation space.

%

\begin{table*}[!t]
    \centering
    \renewcommand{\arraystretch}{1.05}
    \setlength{\tabcolsep}{6pt}
    \caption{Additional comparison of upcycling methods for ViT-B/32. Cluster-aware Upcycling achieves the strongest overall performance across most benchmarks.
    }
    \label{table:supplementary_full_comparison_b32}
    \scalebox{0.9}{
    \begin{tabular}{ccccccccccccc}
    \toprule
    \multicolumn{1}{c}{Model} & 
    \multicolumn{3}{c}{MSCOCO} & 
    \multicolumn{7}{c}{ImageNet-1k} &
    \multicolumn{1}{c}{VTAB} \\
    
    \cmidrule(lr){1-1} \cmidrule(lr){2-4} \cmidrule(lr){5-11} \cmidrule(lr){12-12}
    MoE Init & I$\rightarrow$T & T$\rightarrow$I & Avg. & Val & V2 & A & R & Sketch & ObjNet & Avg. & Natural \\
    \midrule

    Drop-Upcycling ($r$=0.75)~\cite{nakamura2025dropupcycling} & 29.1 & 45.6 & 37.4 & 55.7 & 47.3 & 12.4 & 62.3 & 40.1 & 33.0 & 41.8 & 58.6 \\
    
    Drop-Upcycling ($r$=0.50)~\cite{nakamura2025dropupcycling} & 29.7 & 46.5 & 38.1 & 56.0 & 47.7 & 12.9 & 63.4 & 40.8 & 34.3 & 42.5 & 57.8 \\

    Drop-Upcycling ($r$=0.25)~\cite{nakamura2025dropupcycling} & 30.6 & 47.6 & 39.1 & 56.7 & 48.4 & 13.1 & 63.5 & 41.0 & 35.7 & 43.1 & 57.7 \\

    Sparse Upcycling~\cite{komatsuzaki2023sparse} & 30.8 & {48.0} & {39.4} & 57.1 & {49.1} & 13.8 & 64.3 & 41.8 & 36.0 & 43.7 & 58.0 \\

    CLIP-MoE~\cite{zhang2025clip} & 29.5 & 46.8 & 38.2 & 56.6 & 48.1 & \textbf{14.3} & 64.2 & 41.4 & 35.7 & 43.4 & {58.8} \\

    DeRS-LM~\cite{huang2025ders} & \textbf{31.0} & 47.7 & {39.4} & 56.8 & 48.6 & 13.9 & 64.2 & 41.1 & {36.4} & 43.5 & 58.1 \\

    Drop-SVD & 30.7 & {47.9} & 39.3 & \textbf{57.4} & {48.8} & {14.1} & {64.6} & {42.0} & {36.1} & {43.8} & {58.6} \\

    \textbf{Cluster-aware Upcycling} & \textbf{31.0} & \textbf{48.2} &\textbf{39.6} & {57.3} & \textbf{49.2} & {14.0} & \textbf{65.2} & \textbf{42.3} & \textbf{36.5} & \textbf{44.1} & \textbf{59.1} \\

    \bottomrule
    \end{tabular}}
\end{table*}

\section{Additional Analysis}
\paragraph{Expert utilization}
\label{sec:supplementary_expert_utilization}
%

\begin{figure}[!t]
    \centering
    \begin{subfigure}{\linewidth}
    \centering
    \begin{subfigure}{0.48\linewidth}
        \includegraphics[width=\linewidth]{figures/supplementary/routing_ratios_heatmap_Sparse_Upcycling_vision_gamma0.95.png}
    \end{subfigure}\
    \begin{subfigure}{0.48\linewidth}
        \includegraphics[width=\linewidth]{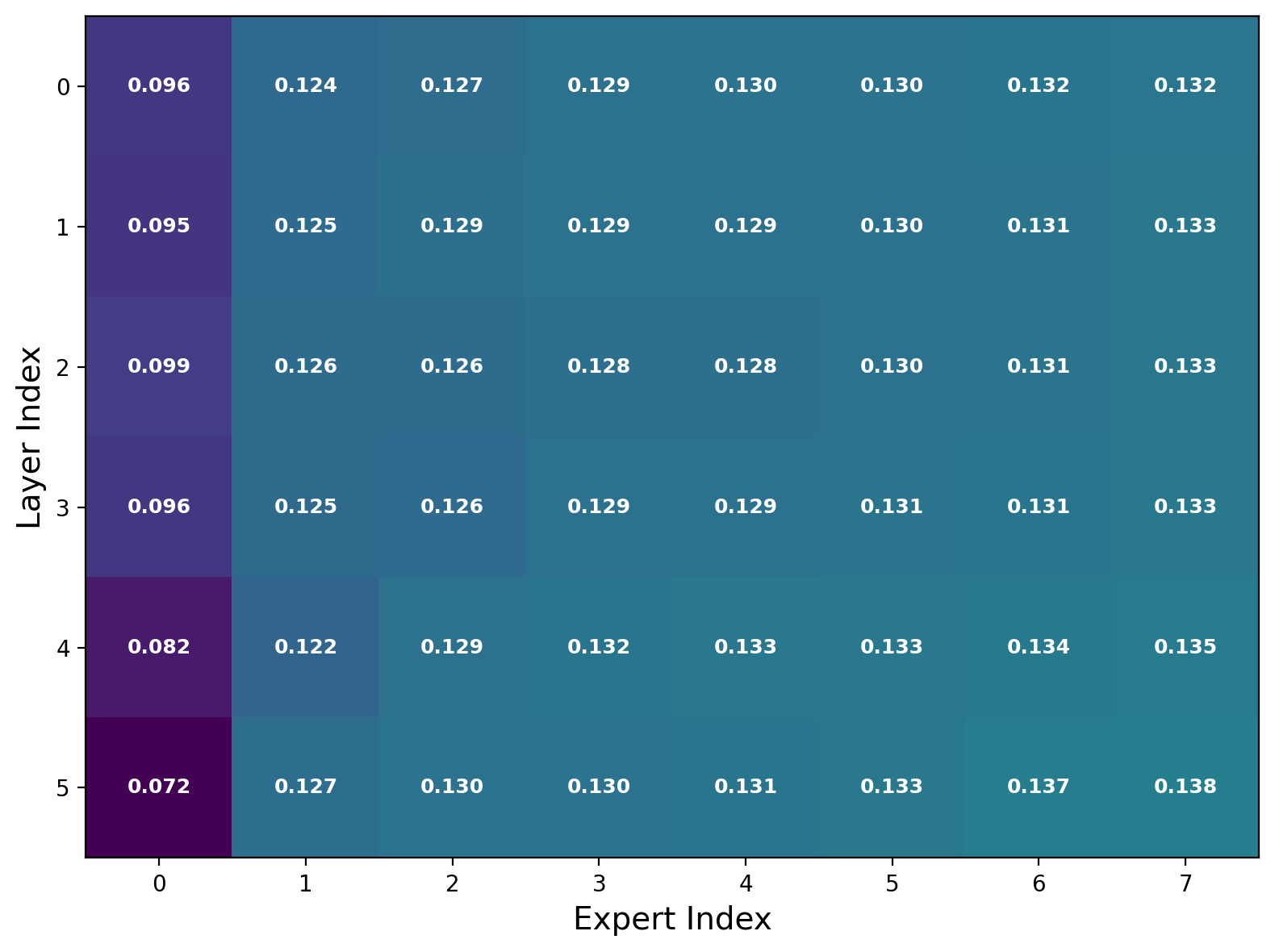}
    \end{subfigure}
    \caption{Sparse Upcycling}
    \end{subfigure}

    \vspace{1mm}
    \begin{subfigure}{\linewidth}
    \centering
    \begin{subfigure}{0.48\linewidth}
        \includegraphics[width=\linewidth]{figures/supplementary/routing_ratios_heatmap_Cluster-aware_Upcycling_vision_gamma0.95.png}
    \end{subfigure}
    \begin{subfigure}{0.48\linewidth}
        \includegraphics[width=\linewidth]{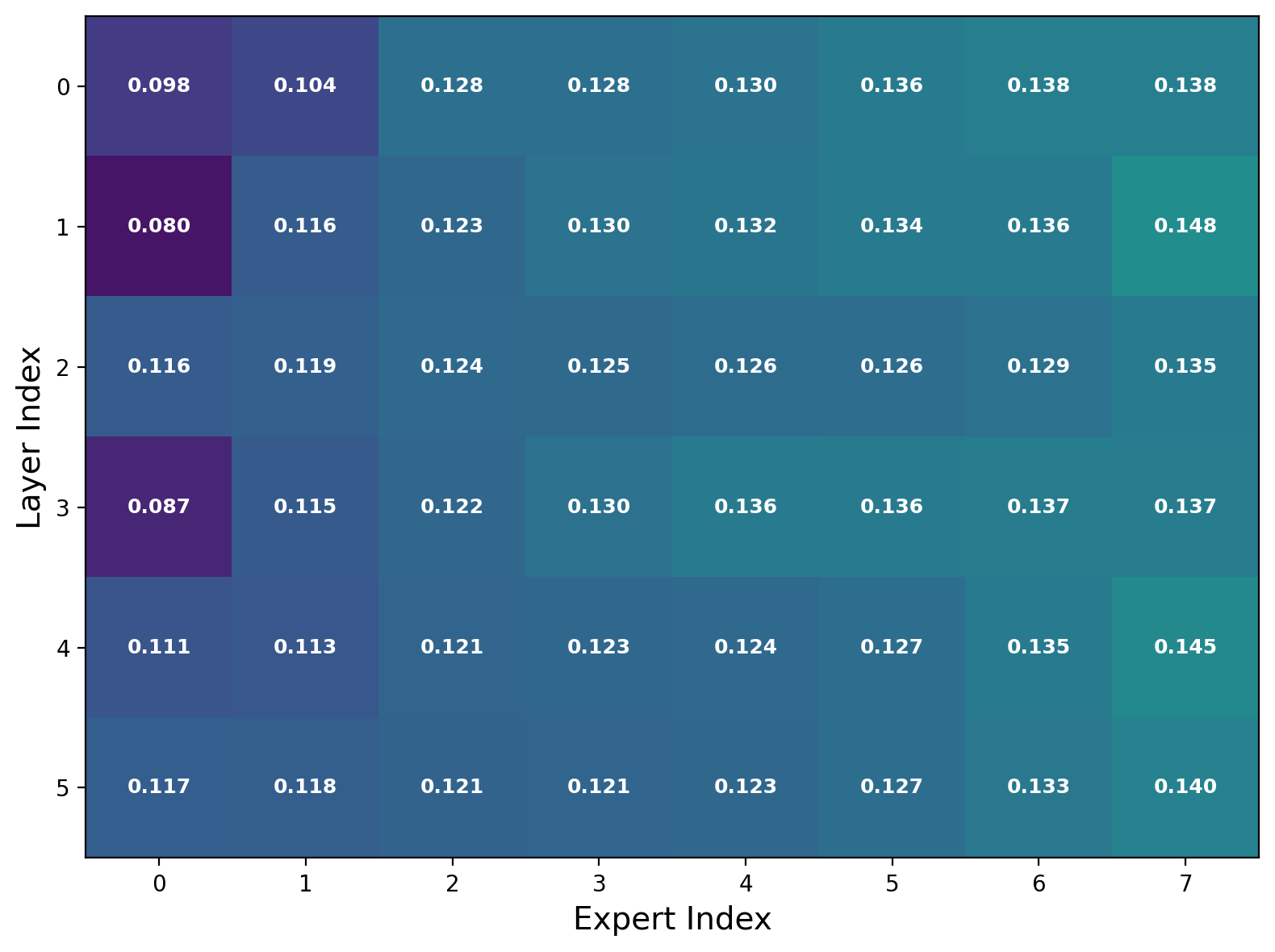}
    \end{subfigure}
    \caption{Cluster-aware Upcycling}
    \end{subfigure}
    \vspace{-4mm}
    \caption{Expert utilization across mixture-of-experts layers for Sparse Upcycling and Cluster-aware Upcycling. Left: Vision towers. Right: Text towers. Best viewed in color.}
    \label{figure:supplementary_analysis_routingpattern_full}
    \vspace{-1mm}
\end{figure}
\begin{figure}[t]
    \centering
    \begin{subfigure}[t]{0.32\linewidth}
        \centering
        \includegraphics[width=\linewidth]{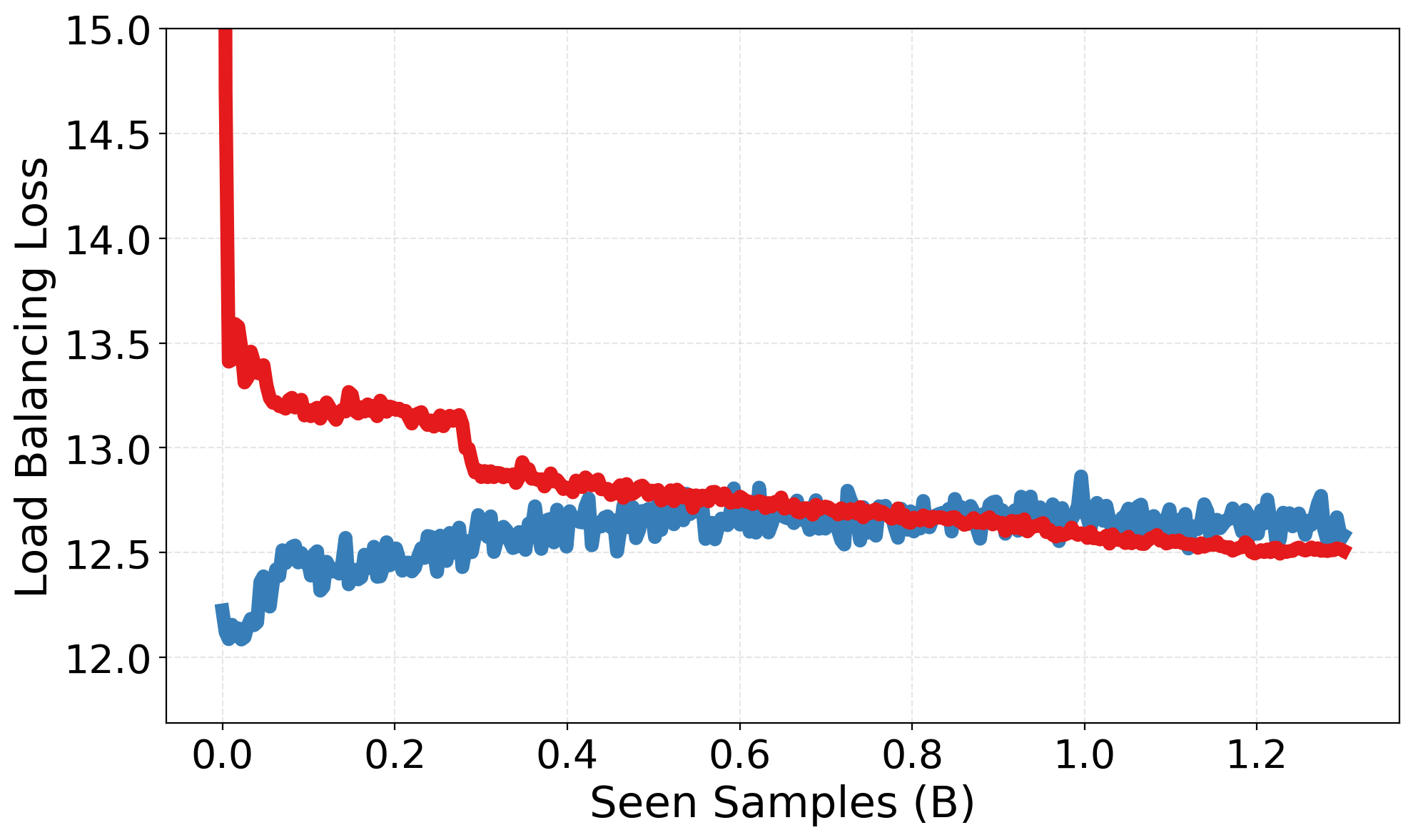}
        \caption{Load-balancing loss}
        \label{fig:supplementary_lb_loss}
    \end{subfigure}
    \hfill
    \begin{subfigure}[t]{0.32\linewidth}
        \centering
        \includegraphics[width=\linewidth]{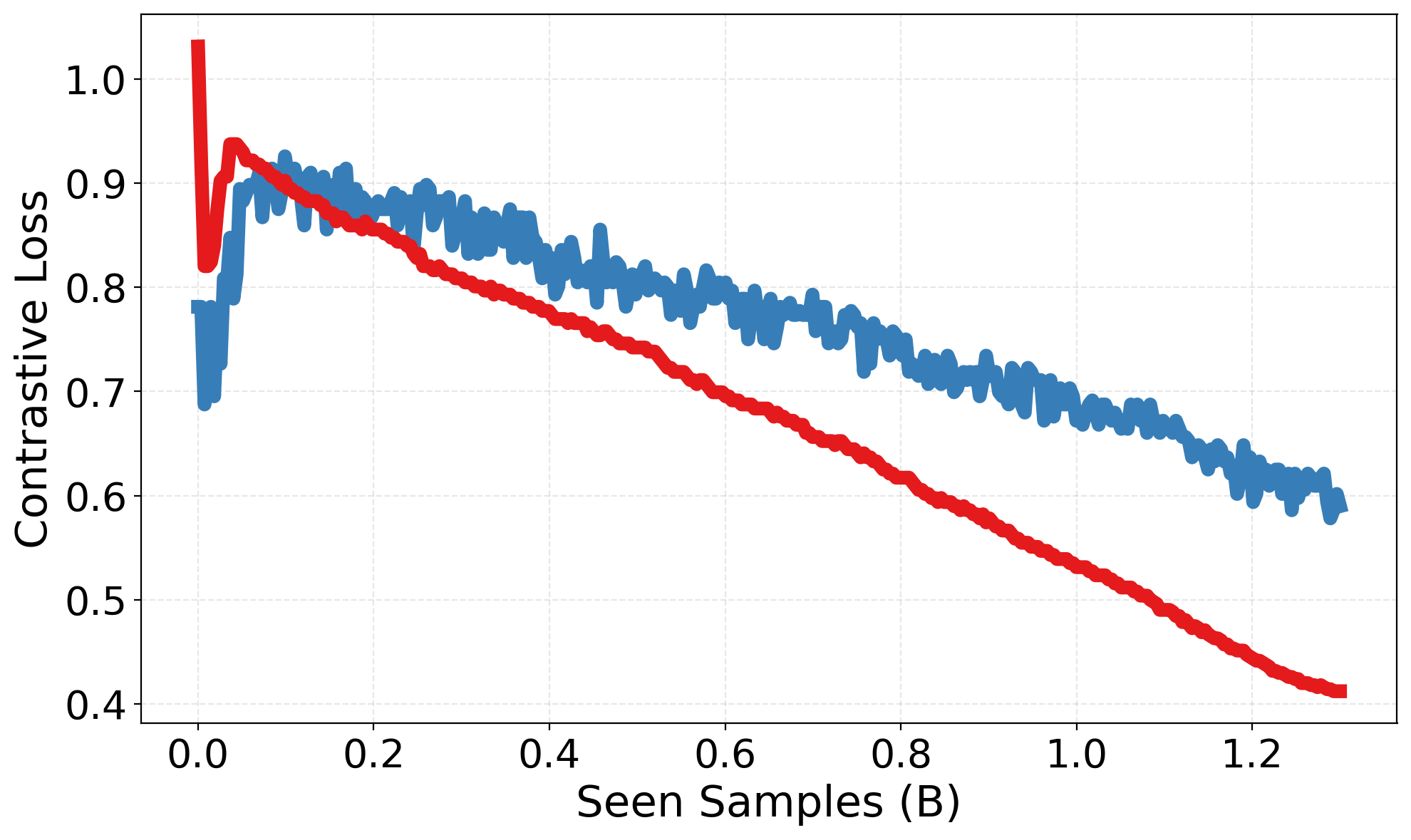}
        \caption{Contrastive loss}
        \label{fig:supplementary_cl_loss}
    \end{subfigure}
    \hfill
    \begin{subfigure}[t]{0.32\linewidth}
        \centering
        \includegraphics[width=\linewidth]{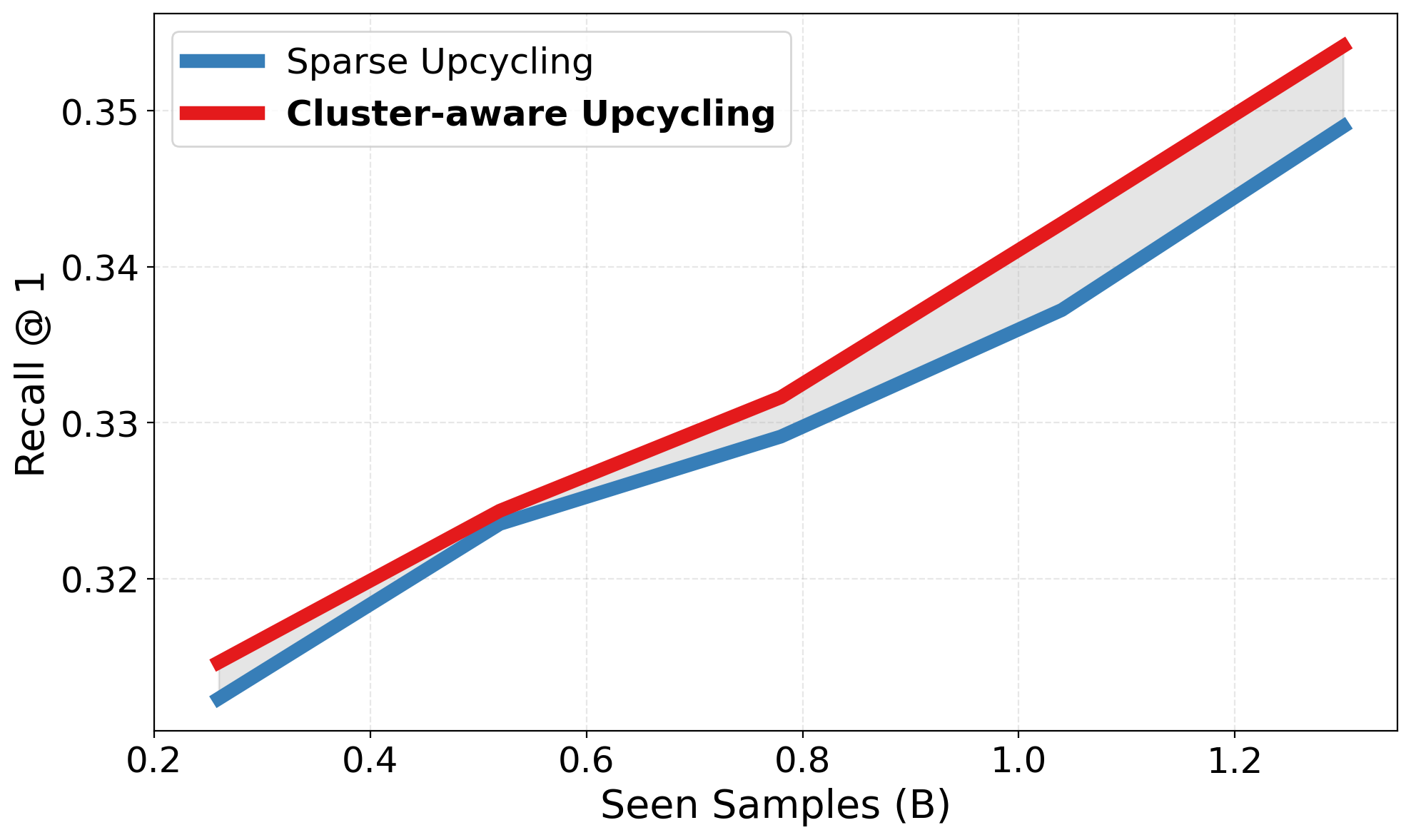}
        \caption{MSCOCO I$\rightarrow$T}
        \label{fig:supplementary_mscoco}
    \end{subfigure}
    \vspace{-1mm}
    \caption{Training dynamics and epoch-wise performance.}
    \label{fig:supplementary_scaling}
    \vspace{-1mm}
\end{figure}

We extend the layer-wise expert utilization analysis of the vision tower shown in Figure~5 to both modalities.
As shown in Figure~\ref{figure:supplementary_analysis_routingpattern_full}, Cluster-aware Upcycling exhibits slightly more diverse expert utilization patterns than Sparse Upcycling, while maintaining balanced expert loads.
Such moderate non-uniformity is expected as specialization emerges, since experts naturally evolve to process different subsets of tokens.
Consistent with these patterns, Figure~\ref{fig:supplementary_lb_loss} shows that the load-balancing loss converges to a value comparable to or lower than that of Sparse Upcycling, confirming well-balanced expert utilization during training.

\paragraph{Scaling behavior}
\label{sec:supple_training_dynamics}
%

We further analyze the scaling behavior and training dynamics of Cluster-aware Upcycling.
As model capacity increases (e.g., ViT-B/32 $\rightarrow$ ViT-B/16), the effectiveness of the proposed method becomes more pronounced, resulting in larger performance gaps between our method and the baselines reported in Table~\ref{table:zeroshot} of the main paper.
Moreover, both the task loss gap and the benchmark performance gap widen as the number of seen samples increases, as illustrated in Figure~\ref{fig:supplementary_cl_loss} and Figure~\ref{fig:supplementary_mscoco}, respectively. 
Our experiments use 5.3B seen samples (4B for dense pretraining and 1.3B for MoE upcycling), whereas typical CLIP pre-training scales to around 13B samples. 
These observations suggest that the proposed method could further benefit from training at larger scales.

\paragraph{Specialization across depth}
%

\begin{figure*}[!t]
    \centering
    \begin{subfigure}[t]{0.24\linewidth}
        \centering
        \includegraphics[width=\linewidth]{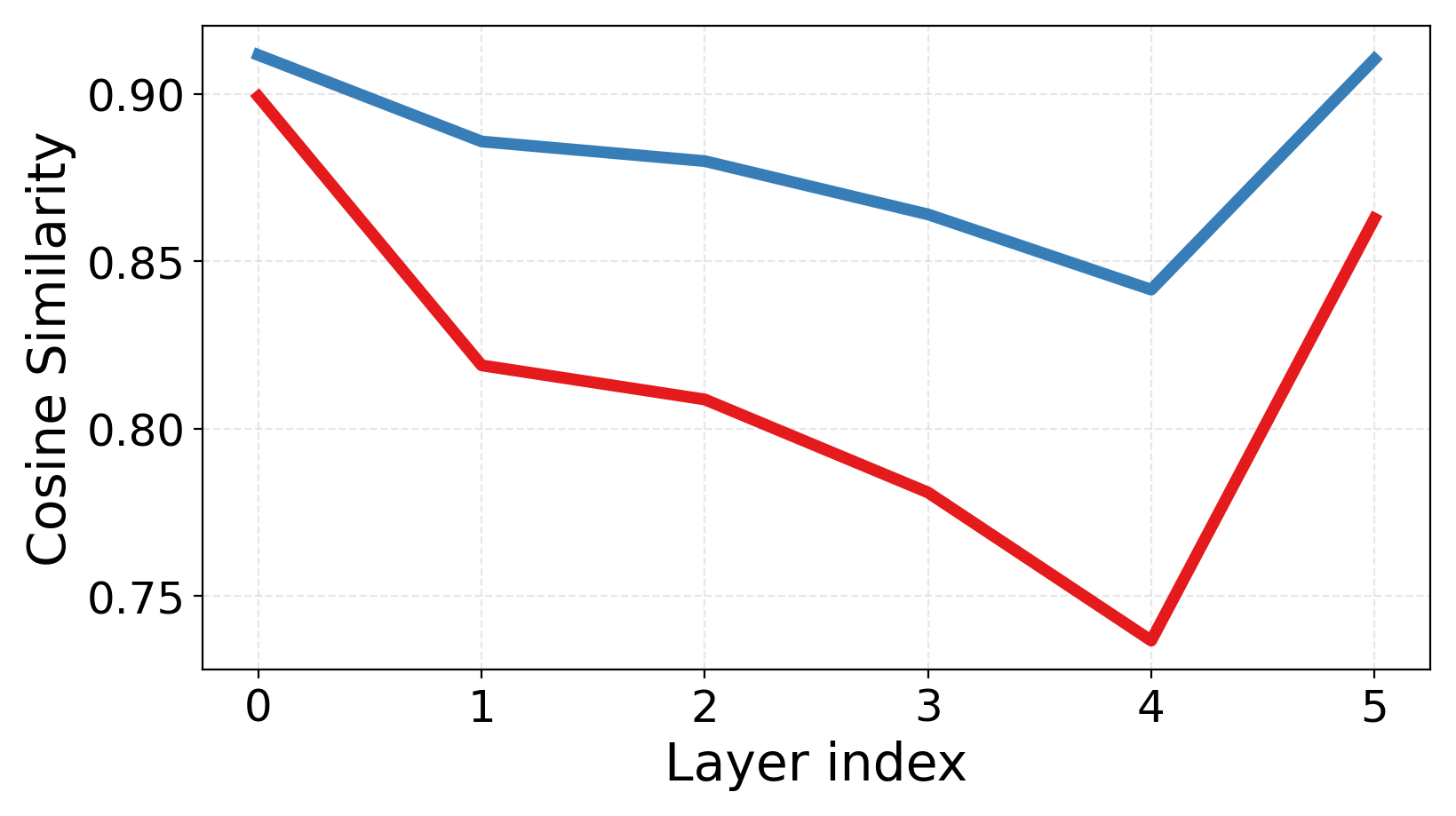}
        \caption{Expert similarity (vision)}
    \end{subfigure}
    \hfill
    \begin{subfigure}[t]{0.24\linewidth}
        \centering
        \includegraphics[width=\linewidth]{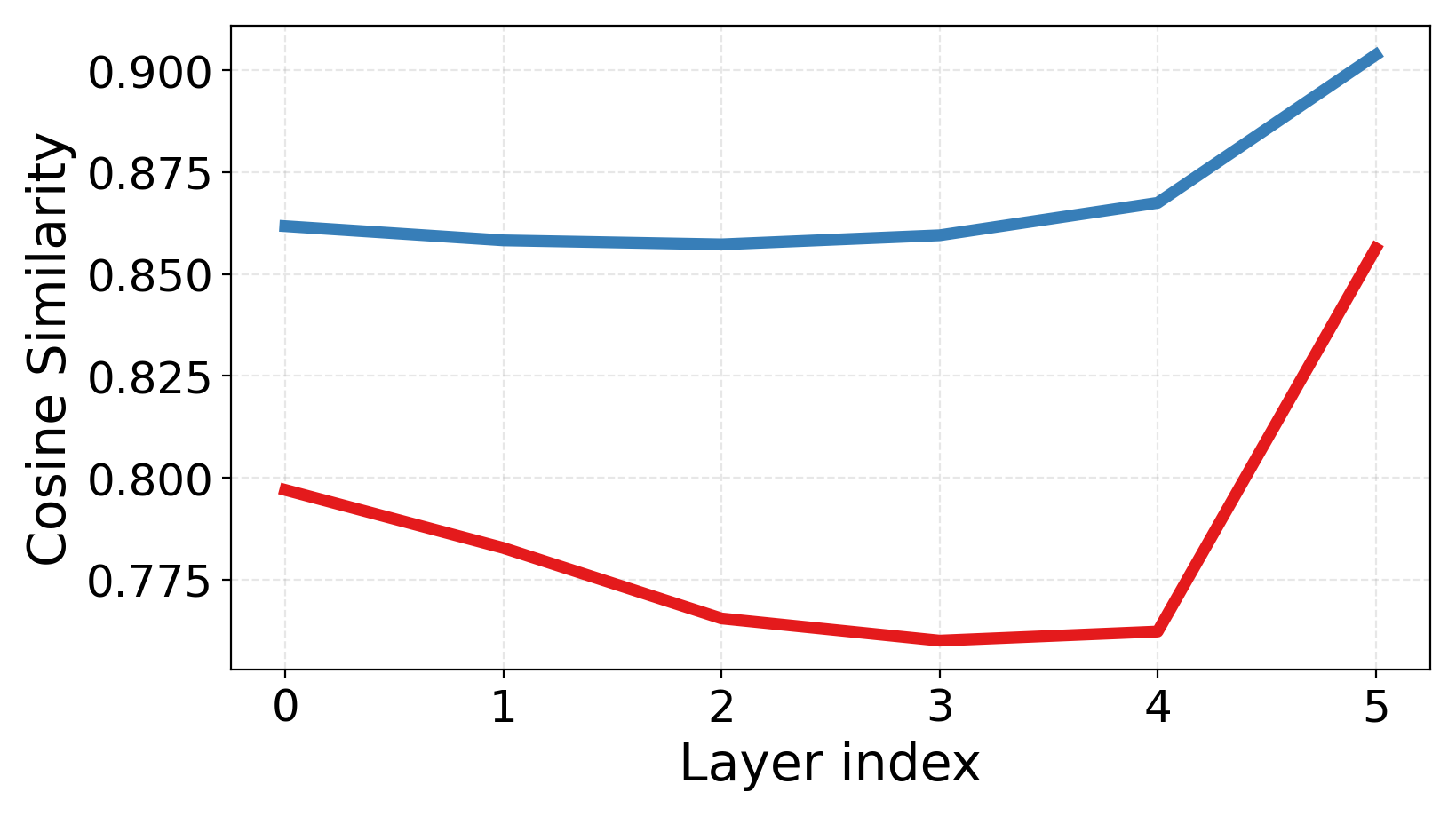}
        \caption{Expert similarity (text)}
    \end{subfigure}
    \hfill
    \begin{subfigure}[t]{0.24\linewidth}
        \centering
        \includegraphics[width=\linewidth]{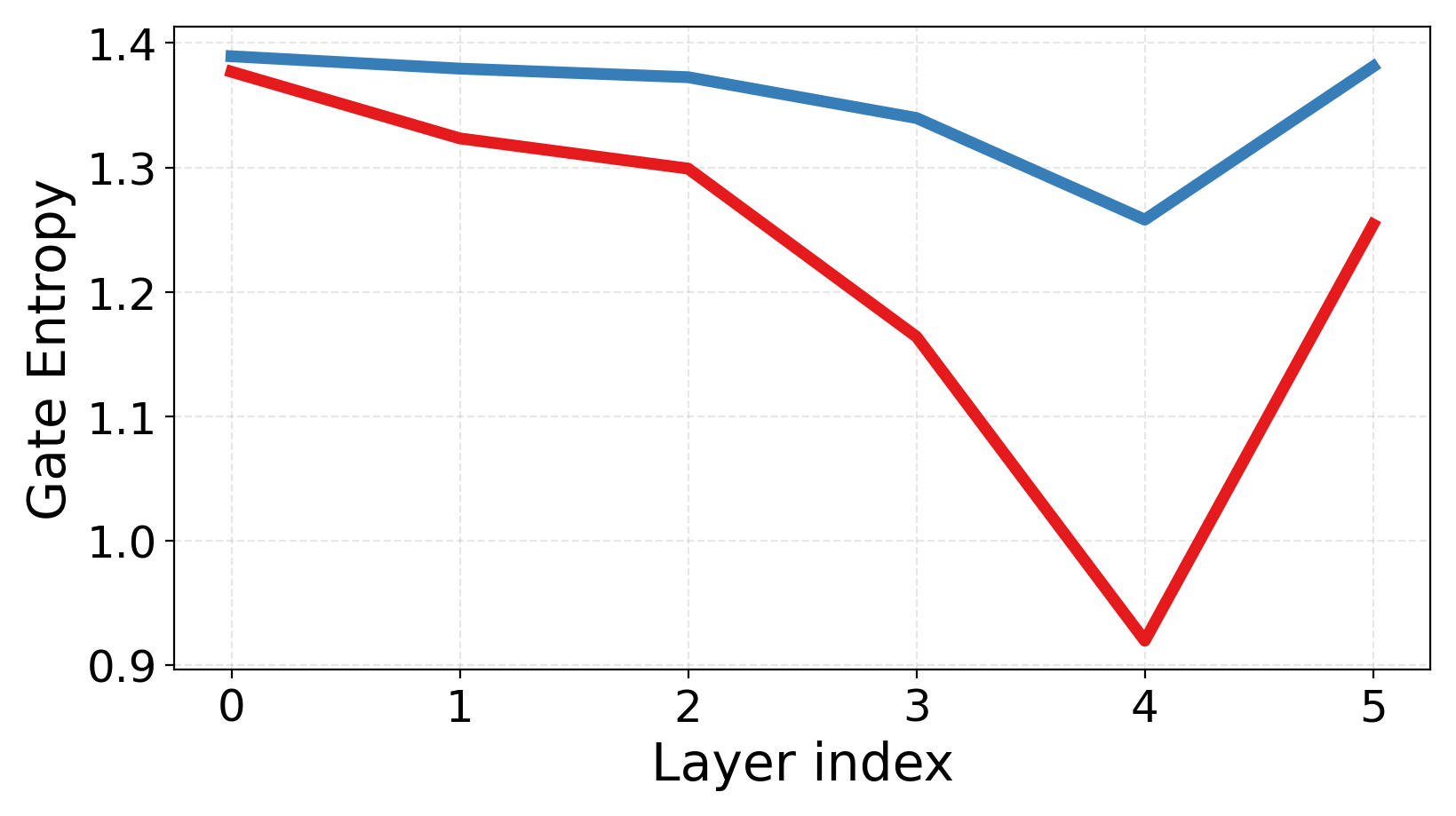}
        \caption{Routing entropy (vision)}
    \end{subfigure}
    \hfill
    \begin{subfigure}[t]{0.24\linewidth}
        \centering
        \includegraphics[width=\linewidth]{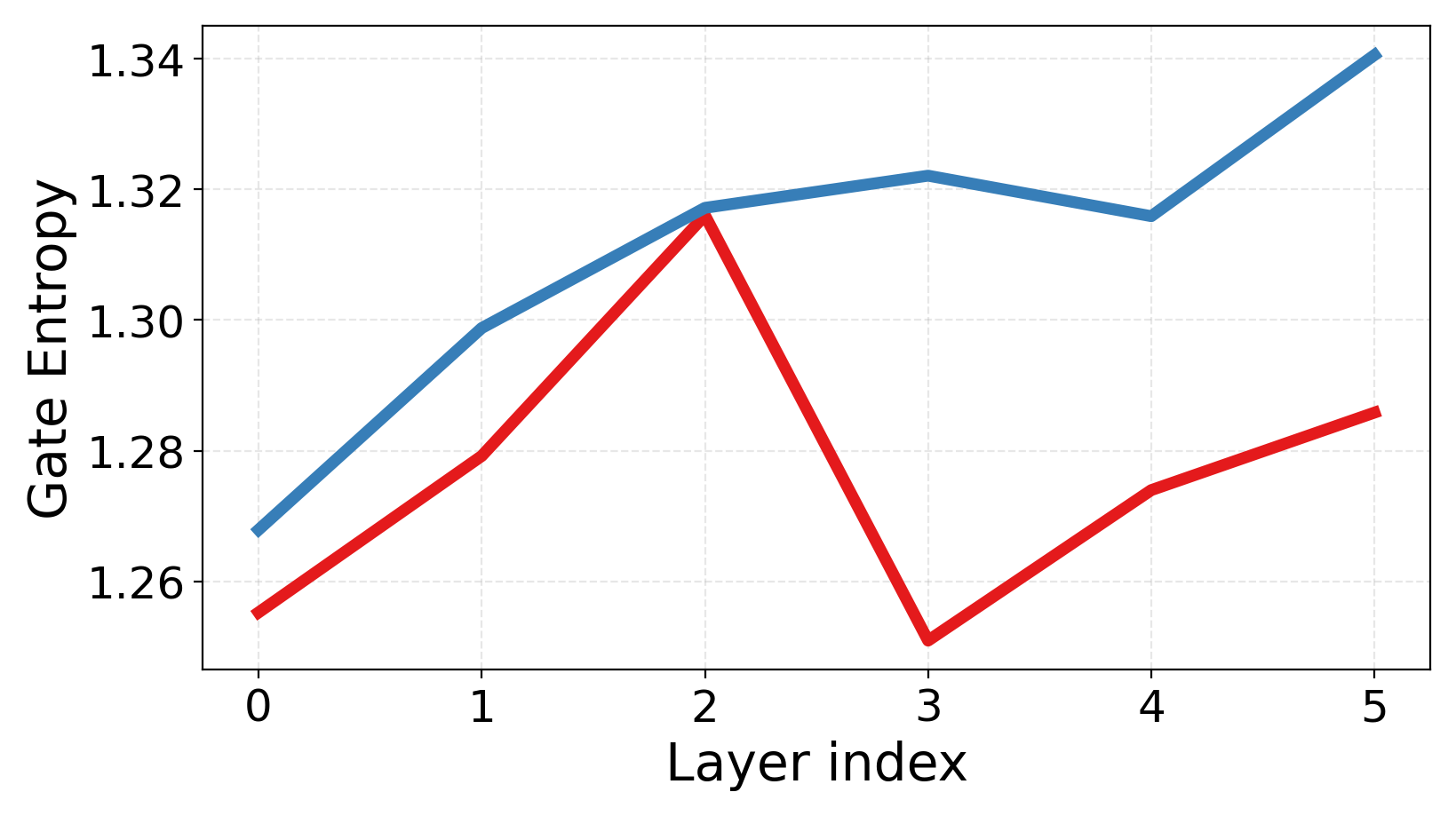}
        \caption{Routing entropy (text)}
    \end{subfigure}
    \vspace{-1mm}
    \caption{Layer-wise expert similarity and routing entropy across MoE layers in the vision and text towers of Sparse Upcycling (blue) and Cluster-aware Upcycling (red).
    }
    \vspace{-1mm}
    \label{figure:analysis_layerwise}
\end{figure*}

To understand how expert specialization evolves across network depth, we illustrate expert similarity and routing entropy across layers in both vision and text towers, in Figure~\ref{figure:analysis_layerwise}.
Across both modalities, expert similarity exhibits a consistent depth-wise pattern: it decreases through most layers, then increases in the final layer, consistent with observations in language MoE models~\cite{lo2025closer}. 
Routing entropy shows a similar trend, with higher entropy in layers where experts are more similar.
However, this pattern is more pronounced in the vision tower, where early layers exhibit higher similarity, possibly due to shared low-level visual features. 
The text tower shows a milder pattern, with less variation across layers.
Despite these modality differences, Cluster-aware Upcycling consistently produces lower expert similarity and routing entropy across all depths, indicating stronger expert differentiation.

\paragraph{Expert-level routing probability}
%

\begin{figure*}[!t]
    \centering
    \begin{subfigure}[t]{0.32\linewidth}
        \centering
        \includegraphics[width=\linewidth]{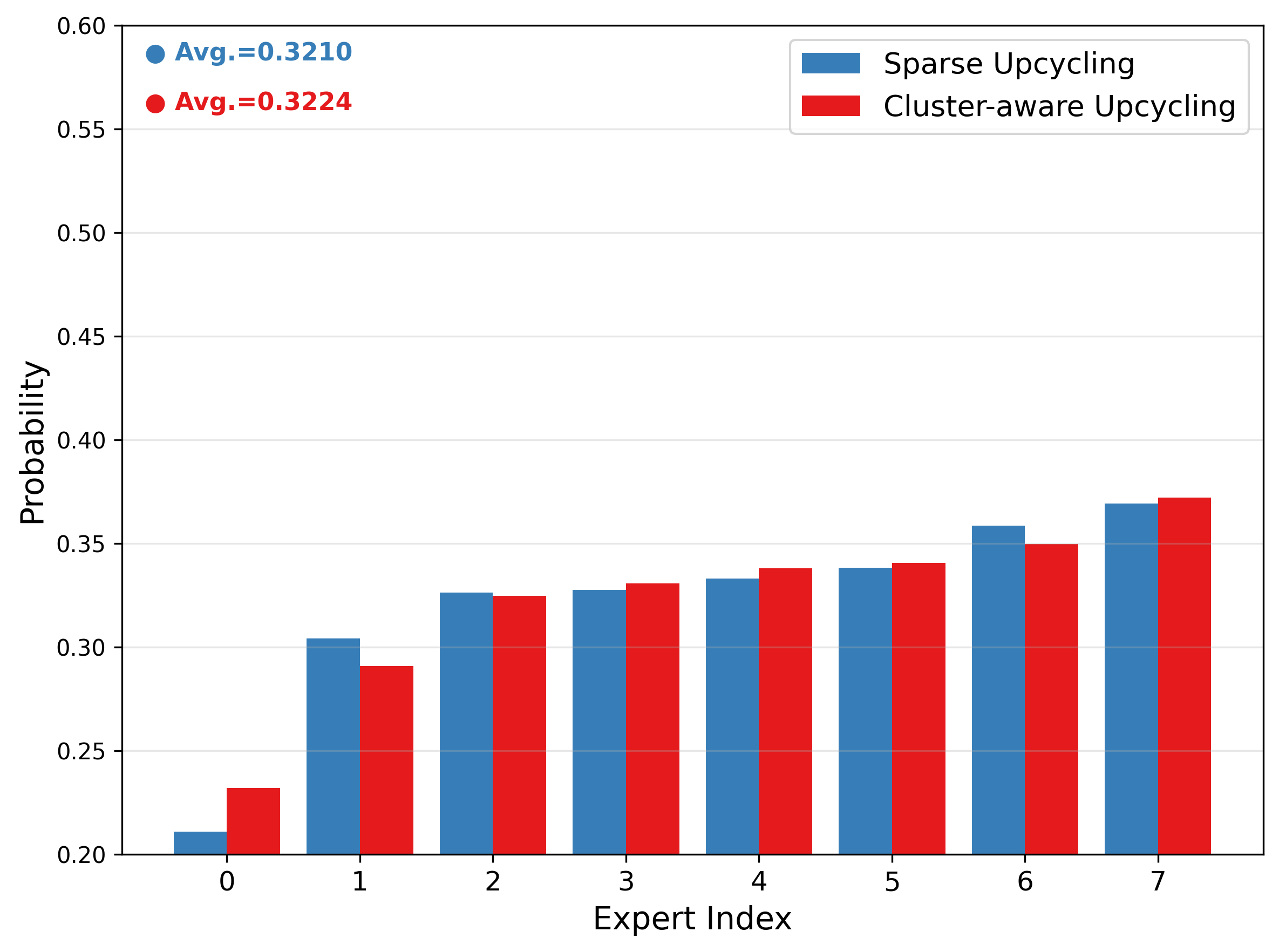}
        \caption{Vision Layer 0}
    \end{subfigure}
    \hfill    
    \begin{subfigure}[t]{0.32\linewidth}
        \centering
        \includegraphics[width=\linewidth]{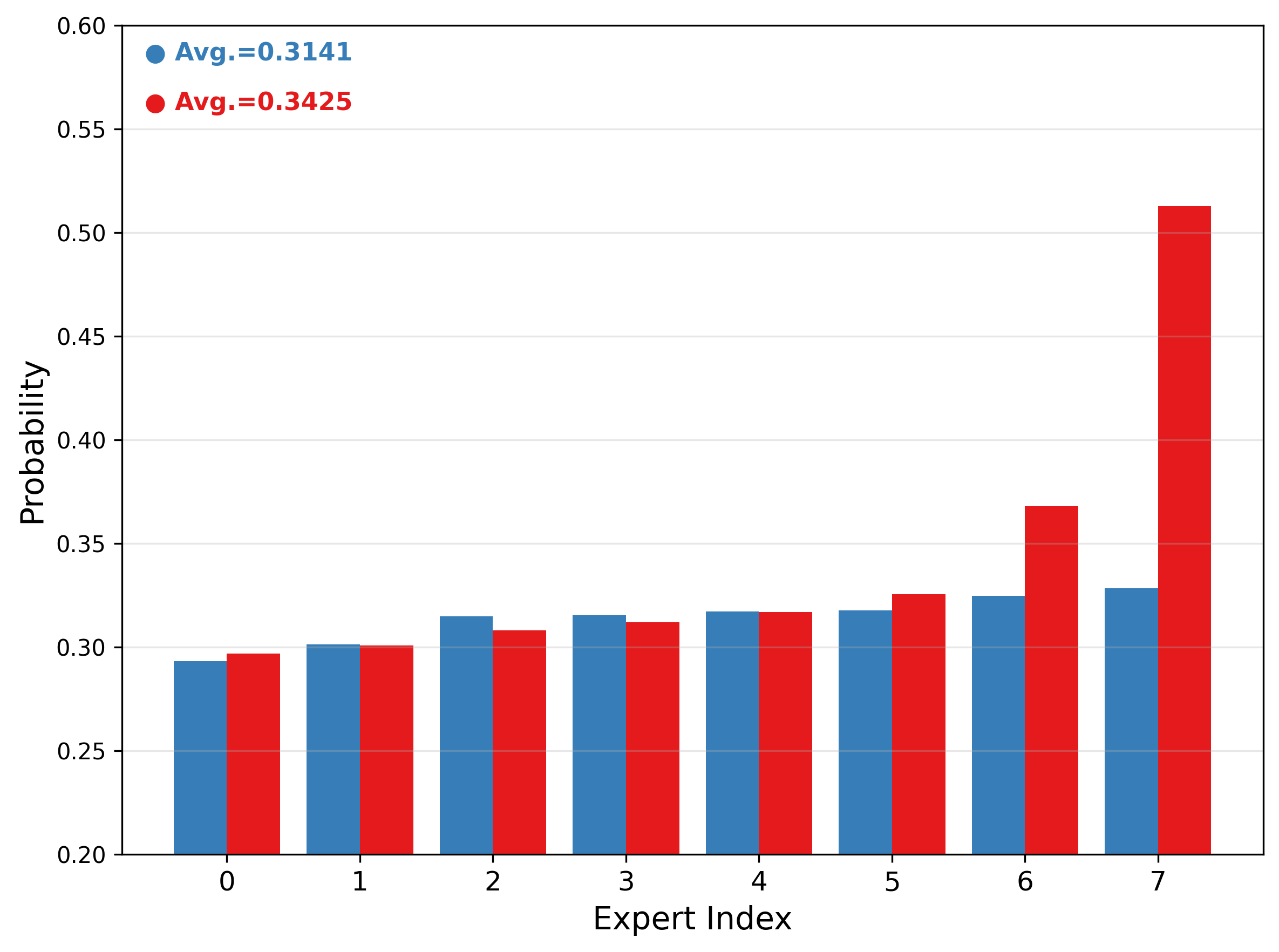}
        \caption{Vision Layer 1}
    \end{subfigure}
    \hfill
    \begin{subfigure}[t]{0.32\linewidth}
        \centering
        \includegraphics[width=\linewidth]{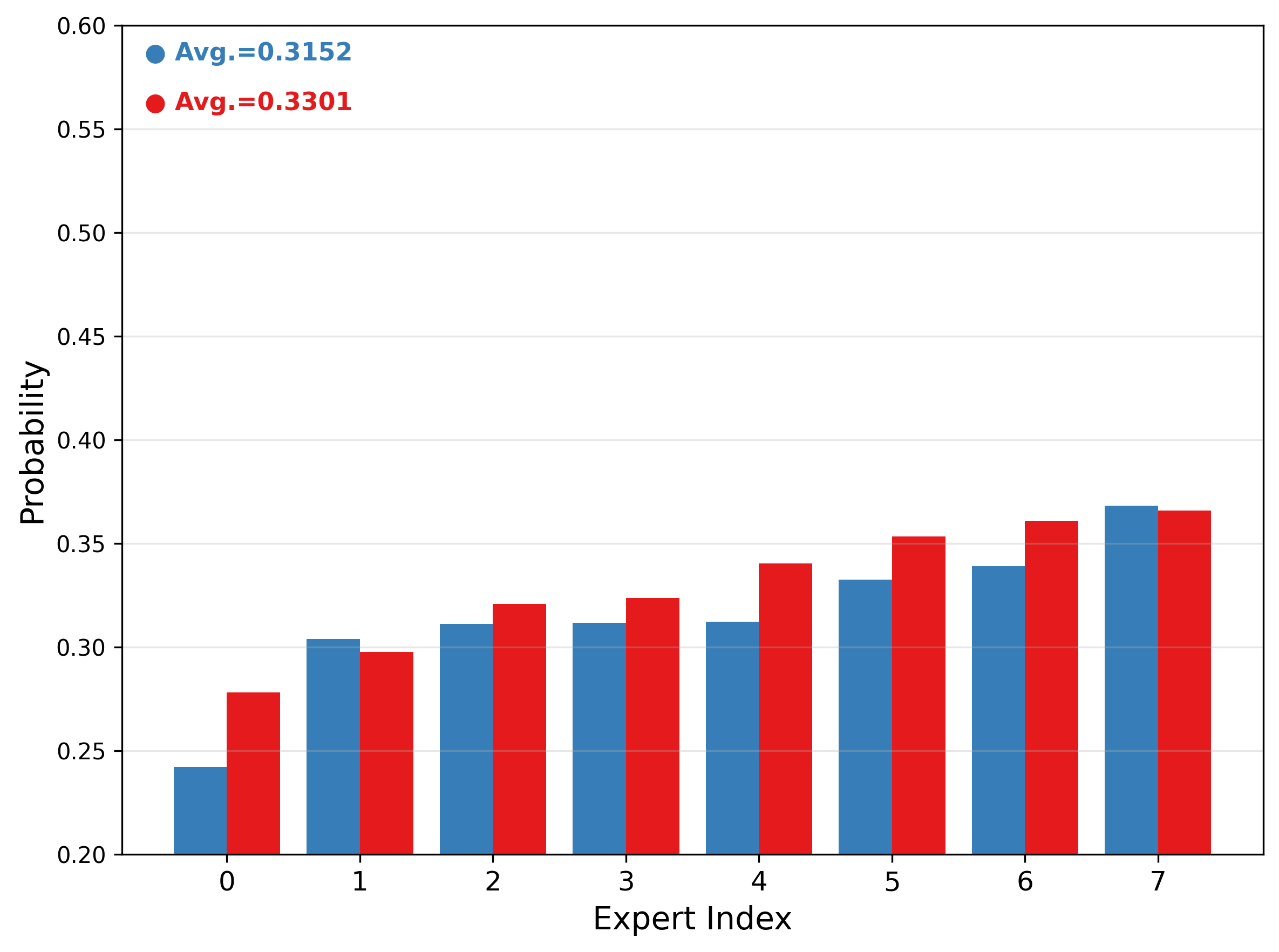}
        \caption{Vision Layer 2}
    \end{subfigure}
    \hfill
    \begin{subfigure}[t]{0.32\linewidth}
        \centering
        \includegraphics[width=\linewidth]{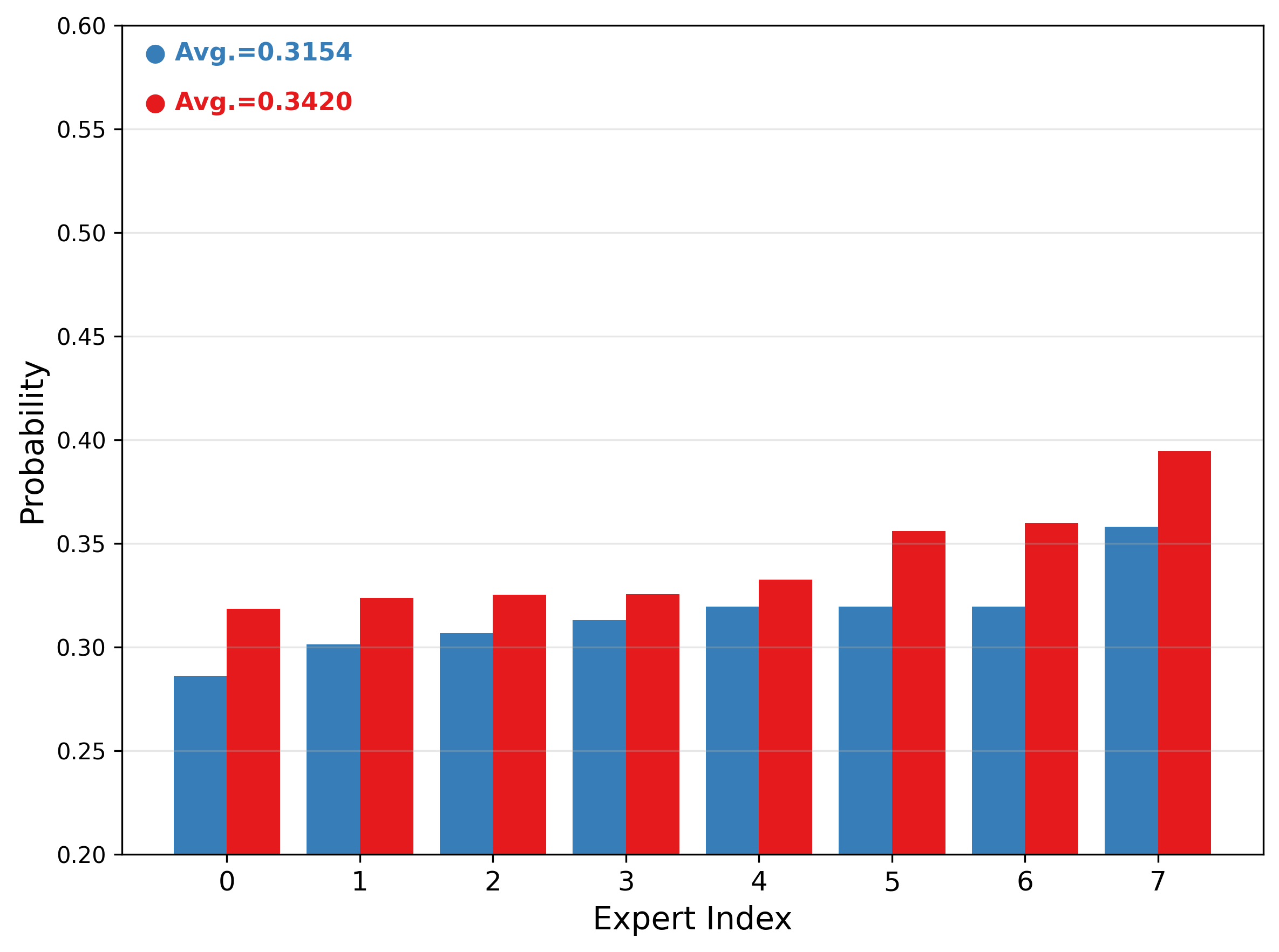}
        \caption{Vision Layer 3}
    \end{subfigure}
    \hfill
    \begin{subfigure}[t]{0.32\linewidth}
        \centering
        \includegraphics[width=\linewidth]{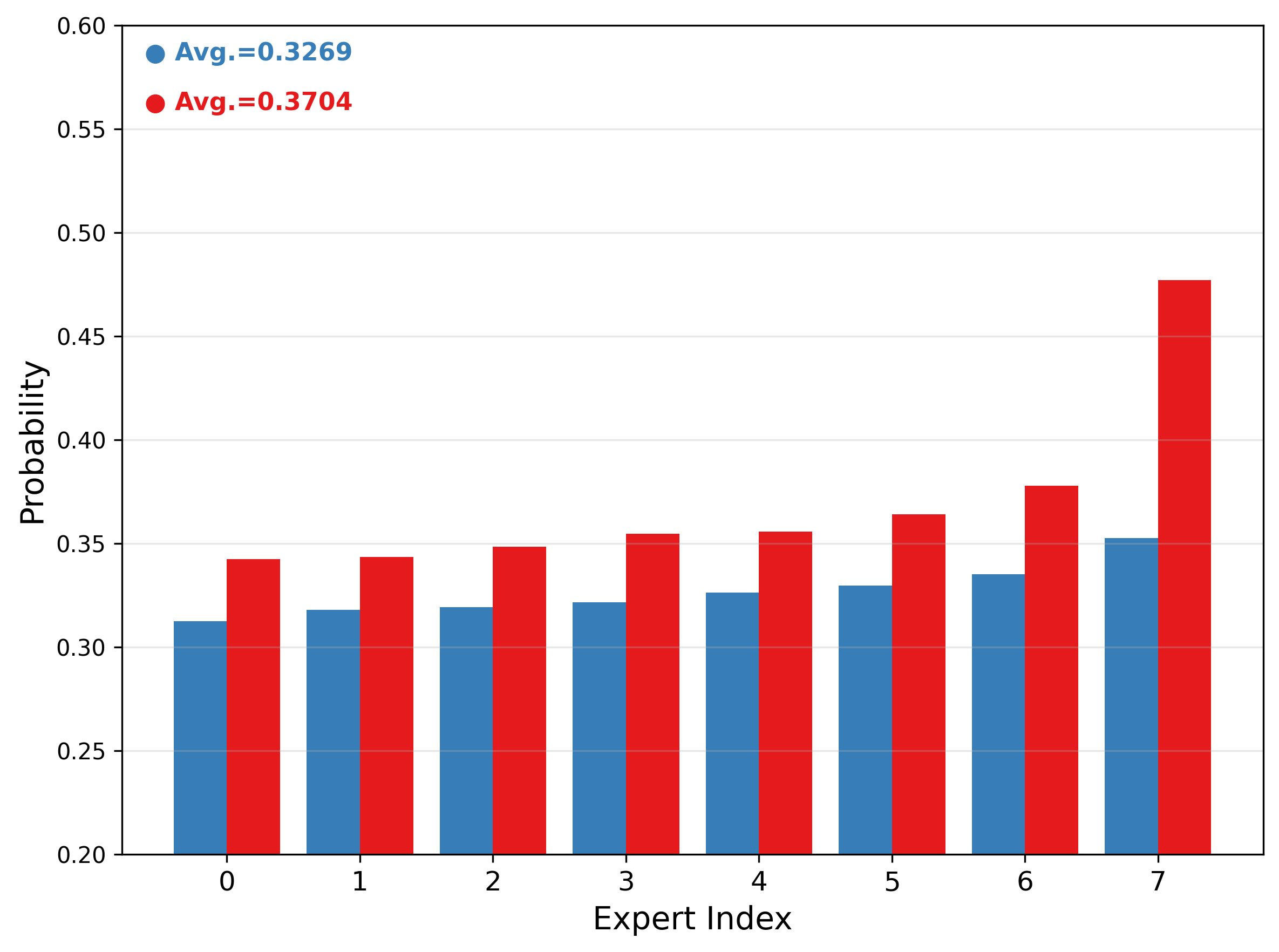}
        \caption{Vision Layer 4}
    \end{subfigure}
    \hfill
    \begin{subfigure}[t]{0.32\linewidth}
        \centering
        \includegraphics[width=\linewidth]{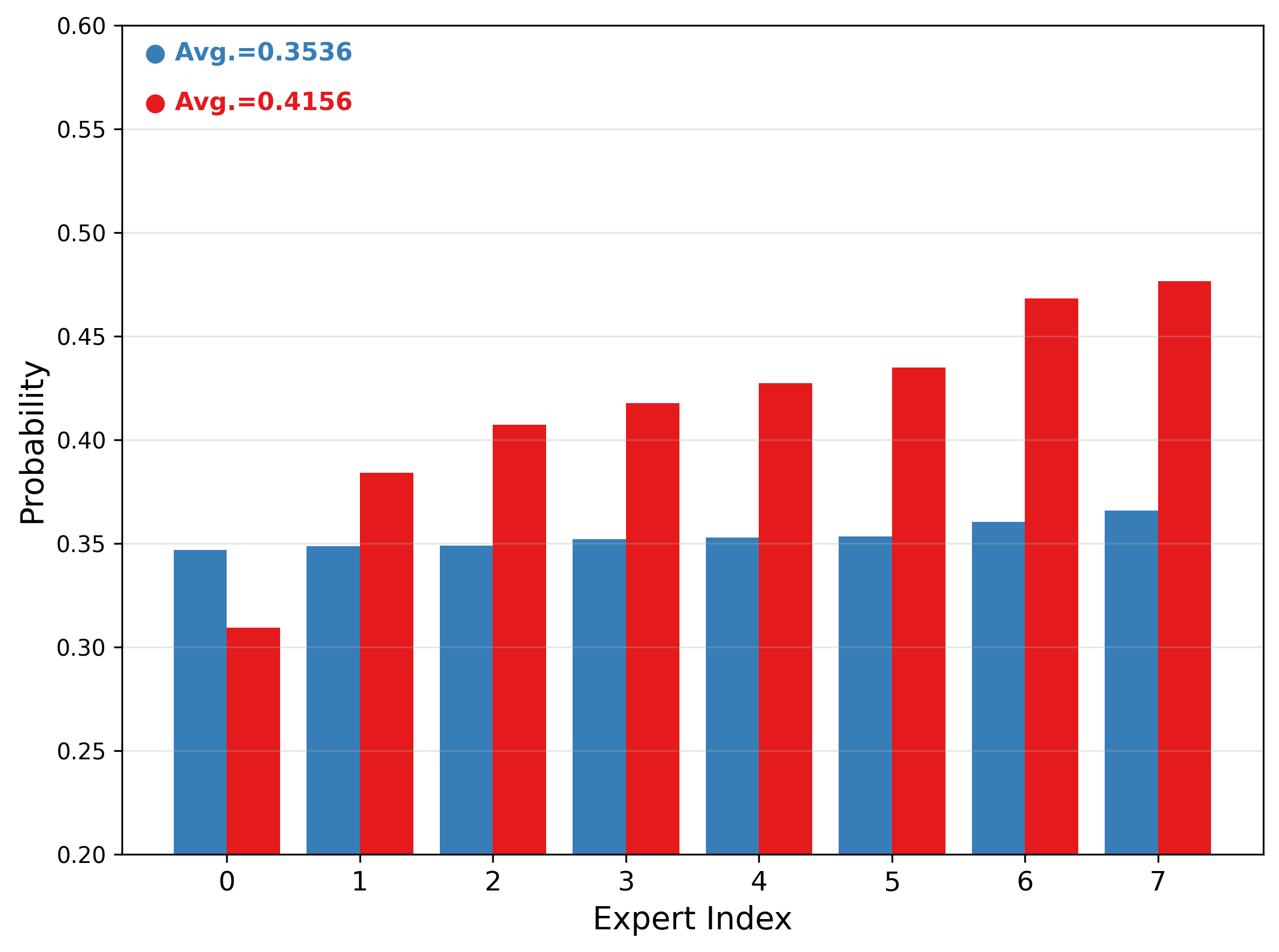}
        \caption{Vision Layer 5}
    \end{subfigure}
    \hfill
    \begin{subfigure}[t]{0.32\linewidth}
        \centering
        \includegraphics[width=\linewidth]{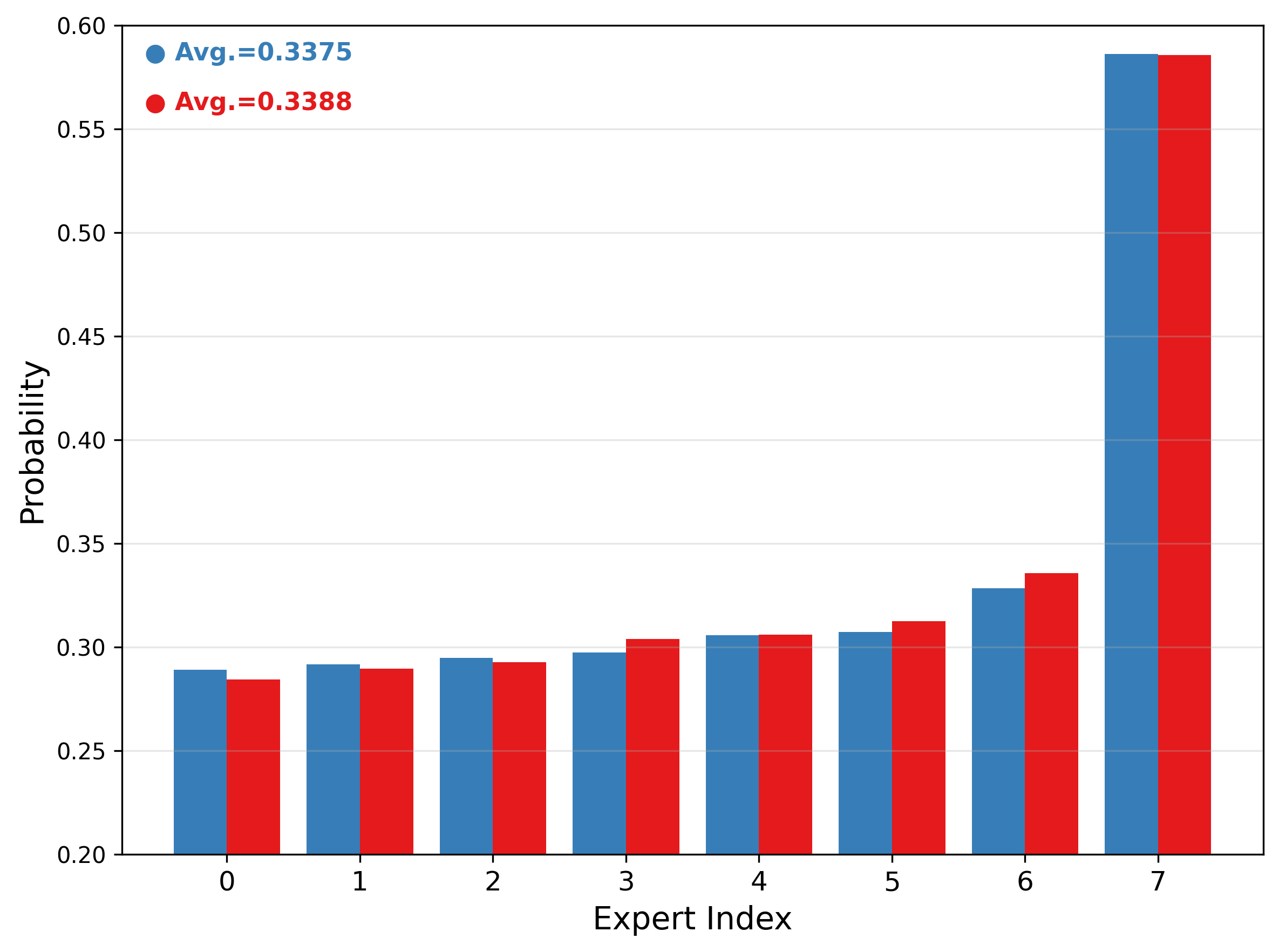}
        \caption{Text Layer 0}
    \end{subfigure}
    \hfill
    \begin{subfigure}[t]{0.32\linewidth}
        \centering
        \includegraphics[width=\linewidth]{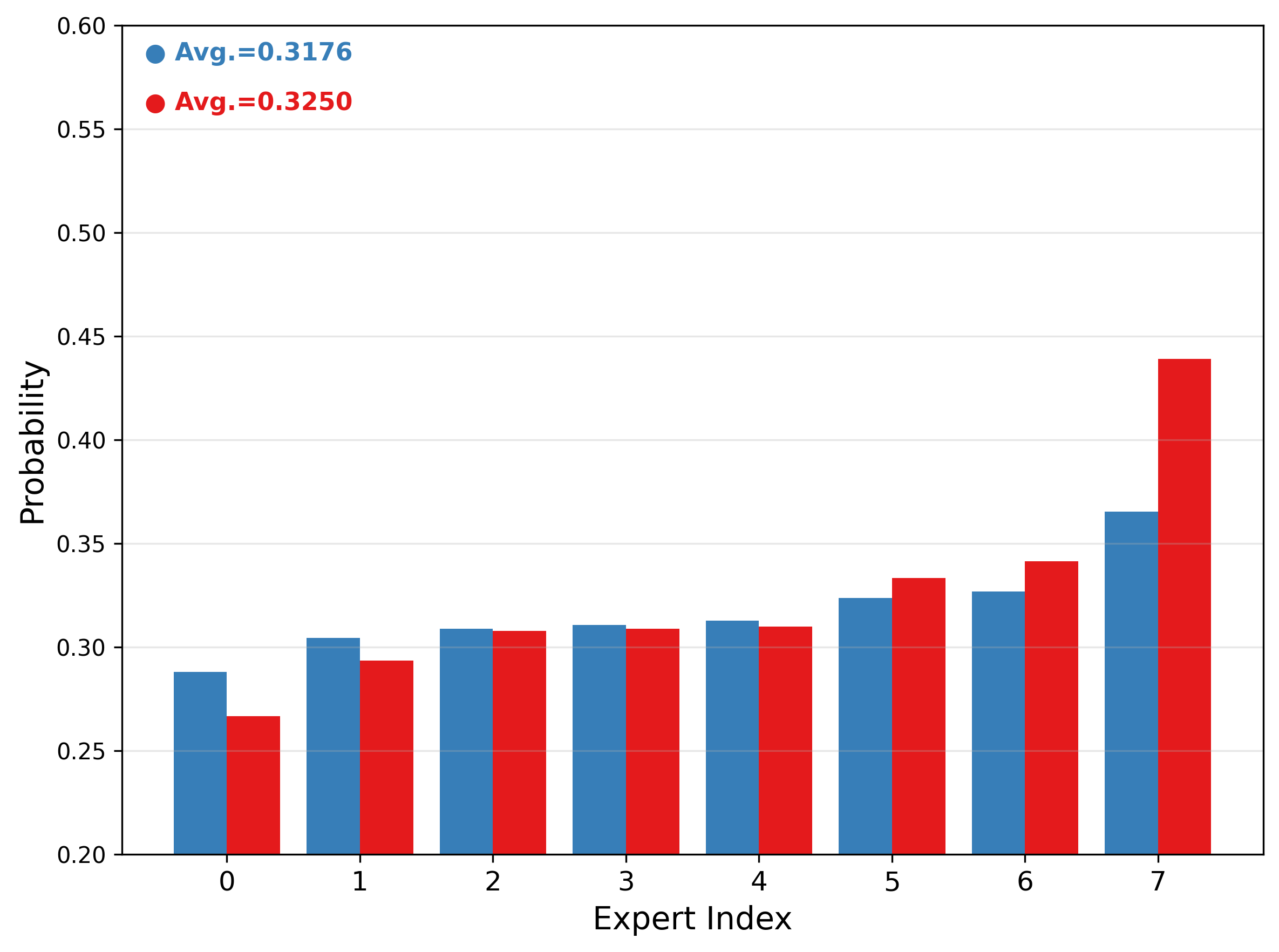}
        \caption{Text Layer 1}
    \end{subfigure}
    \hfill
    \begin{subfigure}[t]{0.32\linewidth}
        \centering
        \includegraphics[width=\linewidth]{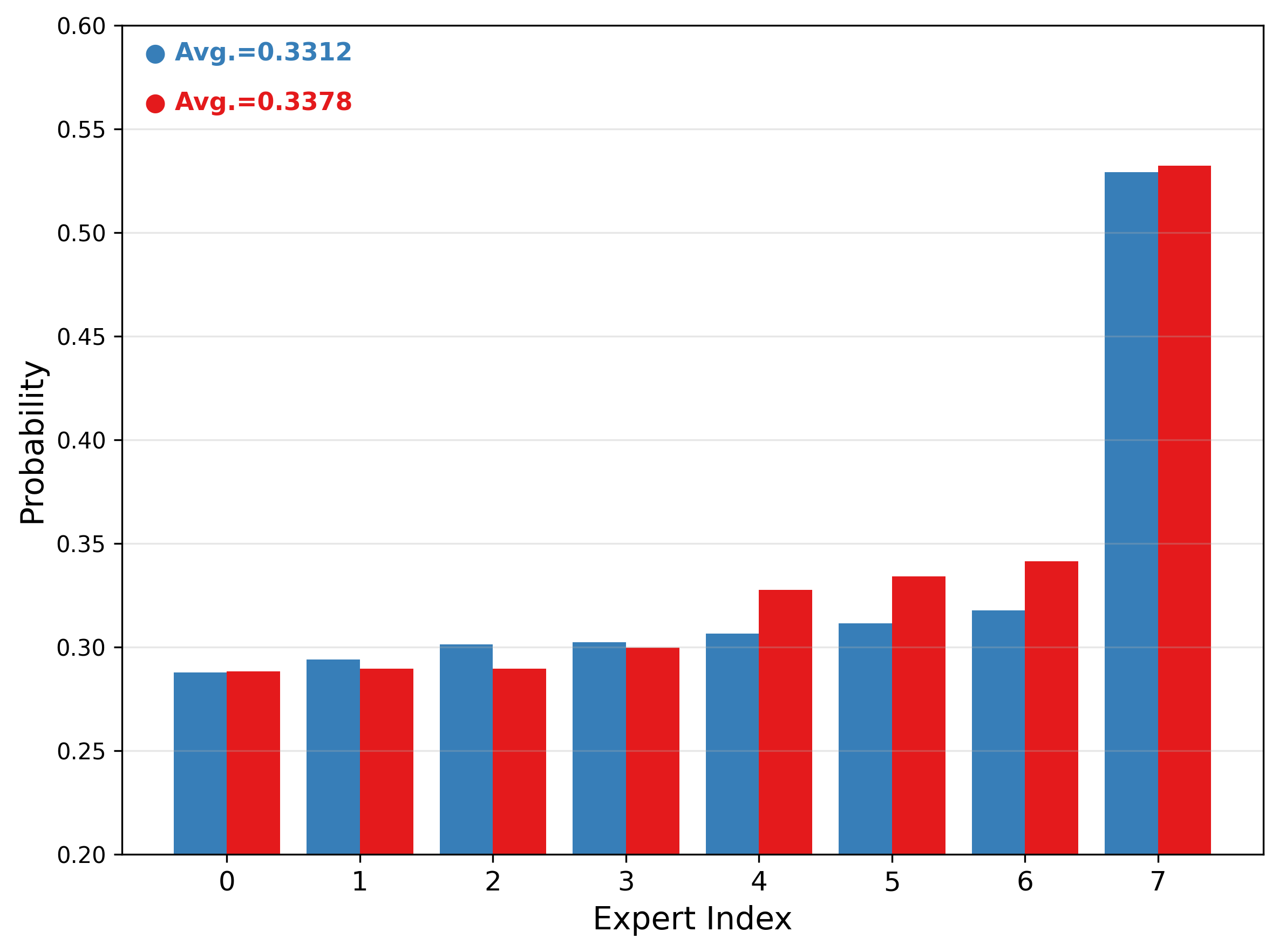}
        \caption{Text Layer 2}
    \end{subfigure}
    \hfill
    \begin{subfigure}[t]{0.32\linewidth}
        \centering
        \includegraphics[width=\linewidth]{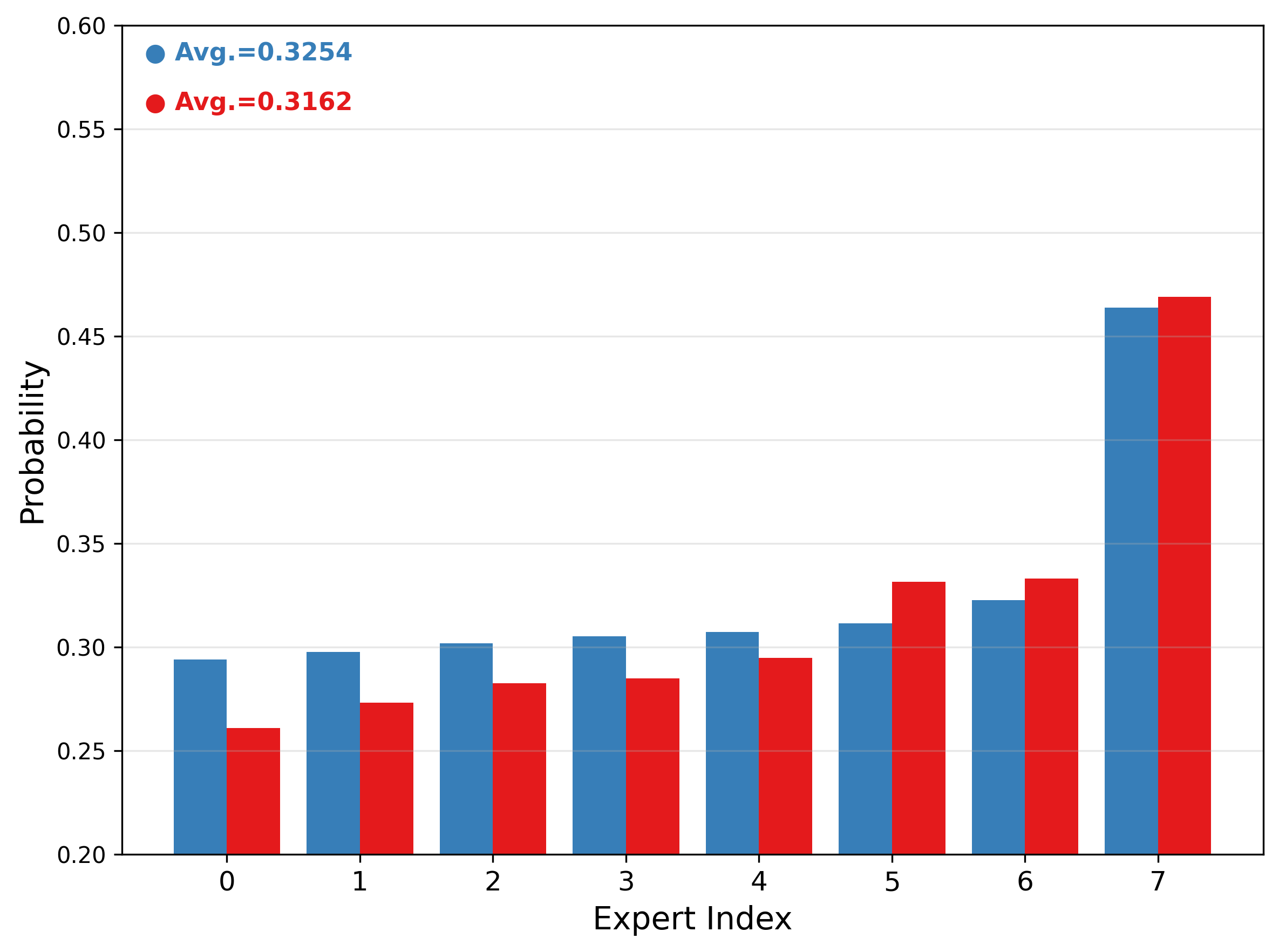}
        \caption{Text Layer 3}
    \end{subfigure}
    \hfill
    \begin{subfigure}[t]{0.32\linewidth}
        \centering
        \includegraphics[width=\linewidth]{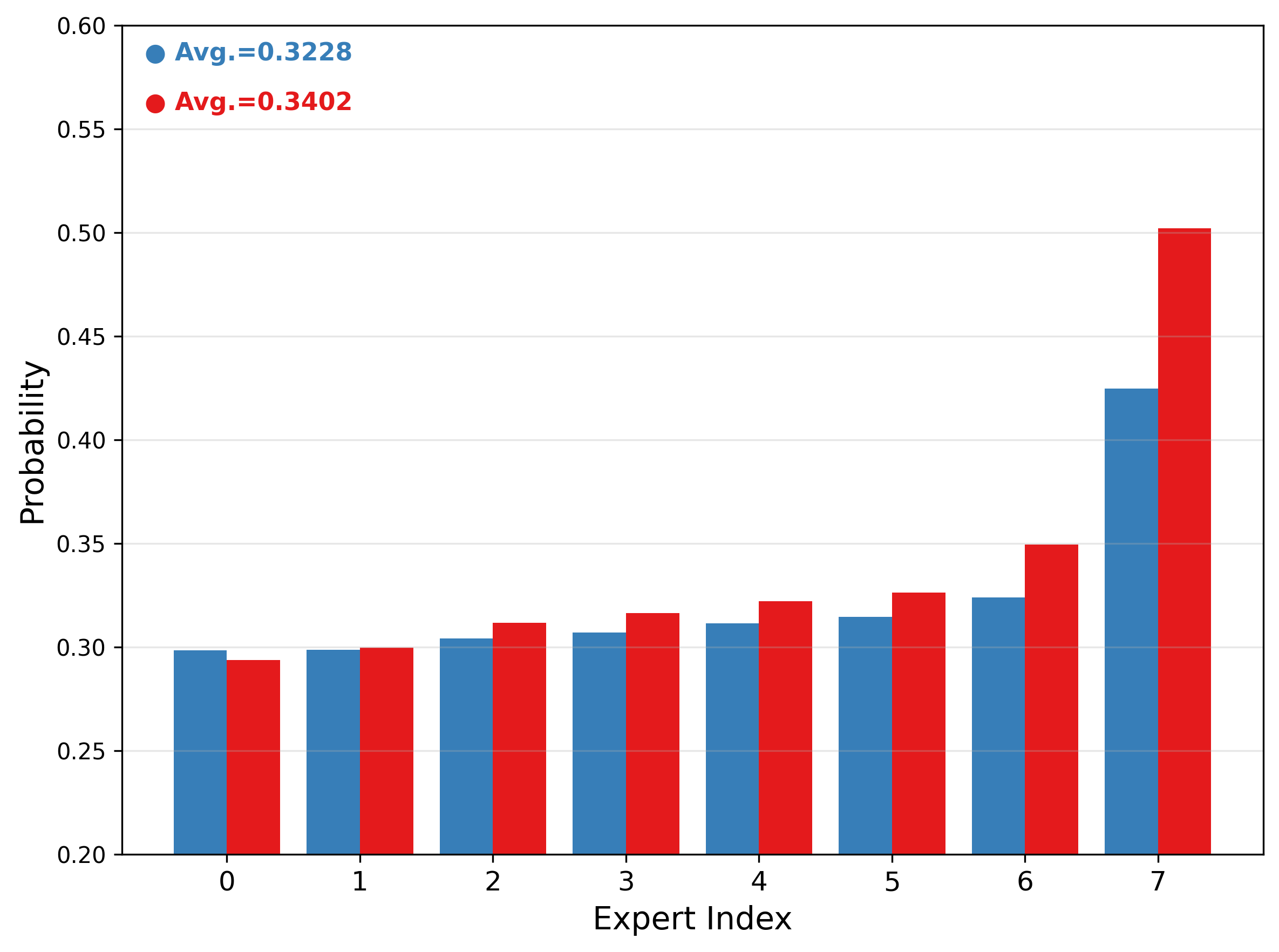}
        \caption{Text Layer 4}
    \end{subfigure}
    \hfill
    \begin{subfigure}[t]{0.32\linewidth}
        \centering
        \includegraphics[width=\linewidth]{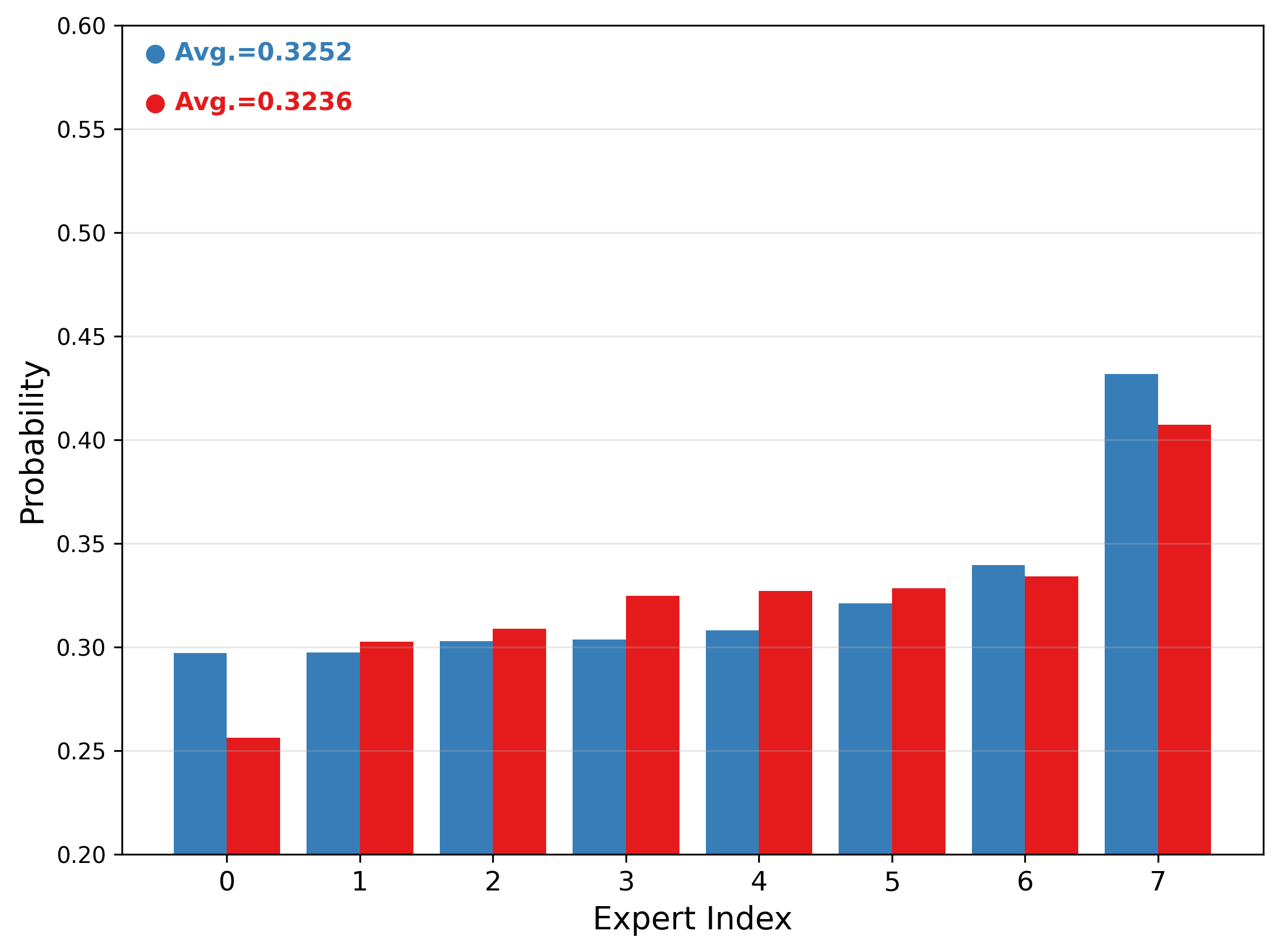}
        \caption{Text Layer 5}
    \end{subfigure}
    \caption{Expert routing probabilities for Sparse Upcycling and Cluster-aware Upcycling.}
    \label{figure:supplementary_analysis_routingmean}
\end{figure*}

To better understand routing behavior, we analyze the average routing probability for each expert, computed over the tokens assigned to that expert.
This provides an expert-level perspective that complements the token-level routing entropy analysis presented in Section~\ref{sec:analysis} of the main paper.
As shown in Figure~\ref{figure:supplementary_analysis_routingmean}, Sparse Upcycling exhibits broadly similar assignment probabilities across experts with top-$2$ routing, whereas Cluster-aware Upcycling shows noticeable variability.
Such variability reflects emerging differences in activation strength across experts and corresponds to a more heterogeneous expert distribution.

\end{document}